# Highlights

## **Data-driven Machinery Fault Diagnosis: A Comprehensive Review**

Dhiraj Neupane,Mohamed Reda Bouadjenek,Richard Dazeley,Sunil Aryal

- This review explores ML algorithms in MFD, including the under-explored RL.
- It assesses diverse MFD methods and extensive dataset usage.
- Covers anomaly detection and physics-informed neural networks for MFD.
- Covers fault detection, classification, and federated learning methods as well.

# Data-driven Machinery Fault Diagnosis: A Comprehensive Review

Dhiraj Neupane*, Mohamed Reda Bouadjenek, Richard Dazeley and Sunil Aryal

*School of IT, Deakin University, WaurnPonds, Geelong, Victoria 3216, Australia*

Abstract

In this era of advanced manufacturing, it's now more crucial than ever to diagnose machine faults as early as possible to guarantee their safe and efficient operation. With the increasing complexity of modern industrial processes, traditional machine health monitoring approaches cannot provide efficient performance. With the massive surge in industrial big data and the advancement in sensing and computational technologies, data-driven machinery fault diagnosis solutions based on machine/deep learning approaches have been used ubiquitously in manufacturing applications. Timely and accurately identifying faulty machine signals is vital in industrial applications for which many relevant solutions have been proposed and are reviewed in many earlier articles. Despite the availability of numerous solutions and reviews on machinery fault diagnosis, existing works often lack several aspects. Most of the available literature has limited applicability in a wide range of manufacturing settings due to their concentration on a particular type of equipment or method of analysis. Additionally, discussions regarding the challenges associated with implementing data-driven approaches, such as dealing with noisy data, selecting appropriate features, and adapting models to accommodate new or unforeseen faults, are often superficial or completely overlooked. Thus, this survey provides a comprehensive review of the articles using different types of machine learning approaches for the detection and diagnosis of various types of machinery faults, highlights their strengths and limitations, provides a review of the methods used for predictive analyses, comprehensively discusses the available machinery fault datasets, introduces future researchers to the possible challenges they have to encounter while using these approaches for fault diagnosis and recommends the probable solutions to mitigate those problems. The future research prospects are also pointed out for a better understanding of the field. We believe that this article will help researchers and contribute to the further development of the field.

## 1. Introduction

### 1.1. Background

Advances in science and technology, coupled with the growth of modern industry, have led to a heightened reliance on machinery, which is often operated under diverse and challenging conditions, including exposure to high humidity levels and excessive loads. These conditions can contribute to machinery failures, with significant impacts such as substantial maintenance costs, decreased production efficiency, financial losses, and, sometimes, the potential for loss of human life [1]. Therefore, the accurate and timely detection of potential machine faults and the implementation of effective maintenance strategies are essential to ensure the continued operation of machines and the safety of human lives.

Both academic and industrial communities have recognized the importance of Machinery Fault Diagnosis (MFD), leading to the development of various diagnostic methods for practical applications [2]. MFD has become an essential part of industrial development and engineering research, playing a vital role in maintaining the safety, reliability, and efficiency of critical machinery in modern industrial processes. Numerous strategies for MFD have been developed by researchers, scientists, and engineers through years of innovative and diligent work. Given the paramount importance of MFD, it is crucial to understand the different techniques of Fault Diagnosis (FD) that have been developed to address various challenges in Machine Health Monitoring (MHM). MHM is a process that involves the continuous evaluation of the overall condition of a machine to ensure optimal performance and prevent unexpected and frequent breakdowns. MFD, on the other hand, is a more focused aspect of MHM, targeting the detection and diagnosis of faults within

*Corresponding author

d.neupane@deakin.edu.au (D. Neupane); reda.bouadjenek@deakin.edu.au (M.R. Bouadjenek); richard.dazeley@deakin.edu.au (R. Dazeley); sunil.aryal@deakin.edu.au (S. Aryal)

ORCID(s): 0000-0001-6548-311X (D. Neupane)

machinery. For the purposes of this review paper, we will use the term MFD to maintain consistency and focus on the detection and diagnosis aspects.

Fault diagnosis is a broad concept that includes various categories, such as fault and/or anomaly detection, fault isolation, fault identification, and fault reconstruction. Fault Detection (FDet) is the first and the easiest step in FD, focusing on identifying whether a fault has occurred. Anomaly Detection (AD) is a proactive approach that identifies unusual patterns in machinery behavior, serving as an early warning system to catch potential faults before they develop into serious issues. Fault identification involves categorizing the fault based on its characteristics, such as whether it is mechanical, electrical, or related to wear and tear. Fault reconstruction, or fault estimation, involves estimating fault amplitude using the idea of redundancy [3]. While FD focuses on identifying and addressing current faults, fault prognosis aims to predict future faults and the Remaining Useful Life (RUL) of machinery components. Prognosis methods are essential for planning maintenance activities before failures occur, thereby reducing downtime and costs [4]. RUL prediction is a key outcome of fault prognosis, providing an estimate of how long a component or system will continue to function before it fails.

To effectively utilize the insights from FD and prognosis, a systematic approach to maintenance is essential. Maintenance strategies can be broadly categorized into three types: Preventive Maintenance (PnM), Predictive Maintenance (PdM), and Reactive Maintenance (RM). [5]. PnM involves regularly scheduled inspections and part replacements to prevent failures before they occur. While effective at reducing unexpected breakdowns, PnM can sometimes result in over-maintenance, which may not be cost-efficient. RM involves taking action only after a failure has occurred. Although RM may seem less resource-intensive initially, it often leads to higher repair costs, increased downtime, and potential safety risks due to unexpected breakdowns [6]. PdM, on the other hand, uses sensor data and analytics to predict when equipment will require repair, allowing maintenance to be scheduled just in time to prevent failures. This approach minimizes downtime and extends the service life of machinery by avoiding unnecessary or premature repairs. Condition-Based Maintenance (CbM) is similar to PdM in that it relies on real-time data to determine the optimal timing for maintenance. As industrial operations move towards Industry 4.0, there is a growing trend toward PdM due to its ability to minimize costs while enhancing productivity and safety. By integrating advanced sensors and real-time data analysis, PdM, often combined with CbM, forms a robust strategy for maintaining machinery health and optimizing performance. In this review article, we focus on PdM and use the term “predictive maintenance" to refer to both PdM and CbM.

As the field of MFD continues to advance and new research emerges continuously (Figure 1 shows the articles published in this field, source Scopus), current research trends and foundational methods are in a state of constant change. Existing reviews focusing on the application of Machine Learning (ML) and Deep Learning (DL) in MFD often deliver a fragmented summary, thus leaving significant gaps in the understanding of this field. While much of the existing literature centres on refining ML/DL models, there has been a significant shift toward understanding how fault information is learned and represented within these models.Numerous studies have surfaced offering improvements in these aspects; however, the literature lacks an overview of the specific areas and the reasons why these methods have been enhanced. Additionally, several review papers tend to provide only brief overviews of numerous research, lacking in-depth exploration. These reviews often offer just one or two sentences per article, which may not offer readers a comprehensive understanding. This survey paper aims to address these issues by offering a comprehensive review of the latest research on MFD using ML and their hybrid, discussing relevant topics, and summarizing the advancements in a more detailed manner.

This paper serves as a comprehensive guide for understanding the application of various ML algorithms in the field of MFD, as well as their respective advantages and disadvantages. Additionally, it addresses the current challenges in intelligent MFD research. Intelligent MFD involves utilizing machine learning techniques to diagnose machine faults. This approach minimizes the need for human intervention and enables automatic detection of machine health states, attracting significant interest over the past two or three decades[7] . Notably, there is a scarcity of review papers dedicated to Reinforcement Learning (RL)-based approaches in MFD. Most existing research papers focus on supervised techniques and briefly touch on un/semi-supervised methods, but do not delve into RL. This gap in the literature has been thoroughly addressed in our review. One of the key distinguishing features of this review is its extensive coverage of available datasets. We provide a comprehensive overview of over 30 datasets, including table detailing relevance, usage, citations, sources, and data types. Additionally, we offer recommendations and discuss challenges for some datasets. While most reviews focus exclusively on vibration data, which remains the most commonly used data type [5], our approach is more comprehensive, encompassing a range of condition monitoring techniques, including thermal imaging, acoustic sensor data, wear debris analysis, oil analysis, and Motor Current

Signature Analysis (MCSA). We have covered each technique, with some discussed in detail and others more briefly. By considering a wider range of condition monitoring techniques, we aim to bridge the gap in literature reviews and provide a comprehensive view of the field. In a nutshell, this review aims to offer a more detailed and comprehensive understanding of the current and future trends in MFD, bridge the gap in literature reviews regarding RL-based works, and provide insights into various condition monitoring techniques beyond just vibration signals.

### 1.2. Motivation

Given the widespread use of both DL and RL-based methods in MFD in recent years, a well-structured review of the relevant literature has become essential. Such a review would support future research in the field by providing a comprehensive summary of past studies, laying the groundwork for more in-depth explorations. There are some existing reviews summarizing the related literature from different perspectives. Zhang et al. [8], Neupane et al. [1], Hoang et al. [9], Mushtaq et al. [10], Singh et al. [11], Tang et al. [12], Sunal et al. [13], Kumar et al. [14], AlShorman et al. [15], Helbing et al. [16], Stetco et al. [17] have summarized and overviewed the ML/DL-based approaches for MFD of a particular component or equipment like rolling element bearings, gears, pumps or induction motors. Soother et al. [18] and Tang et al. [19] highlighted the importance of data processing for condition monitoring and analyzed the existing methods for data processing for DL-based MFD. Also, Zhang et al. [20] discussed the small and imbalanced data in MFD and provided the data augmentation-based, classifier design-based, and feature learning-based techniques to mitigate the issues. Similarly, Li et. al [21], Zheng et al. [22], Li et al. [2], Liu et al. [23], Hakim et. al [24] provided the review of transfer learning and domain adaptation for the machinery fault detection. Ruan et al. [25] and Pan et al. [26] focused on the Generative Adversarial Network (GAN)-based approaches for data augmentation, Anomaly Detection (AD) and fault detection and classification for machine data. Moreover, Zhu et al. [27] reviewed the applications of Recurrent Neural Network (RNN) to mechanical fault diagnosis. Yang et al. [28] and Qian et al. [29] provided a comprehensive review of Autoencoder (AE)-based approaches for industrial applications and pointed out the challenges and prospects of AE-based MFD research. Moreover, the basic structure and principles of Convolutional Neural Networks (CNNs) are discussed by Tang et al. [19] and Jiao et al. [30], focusing on analyses and summary of the applications of CNNs for fault diagnosis in rotating machinery. At the time of writing this manuscript, we couldn't locate any reviews specifically addressing RL for MFD. Thus, this article provides a thorough review of the latest advancements in MFD, employing Traditional Machine Learning (TML), DL, RL, signal-processing (SP), AD approaches, transformers, and Physics Informed Neural Networks (PINNs). Table 1 provides the comparison of our article with the other available reviews.

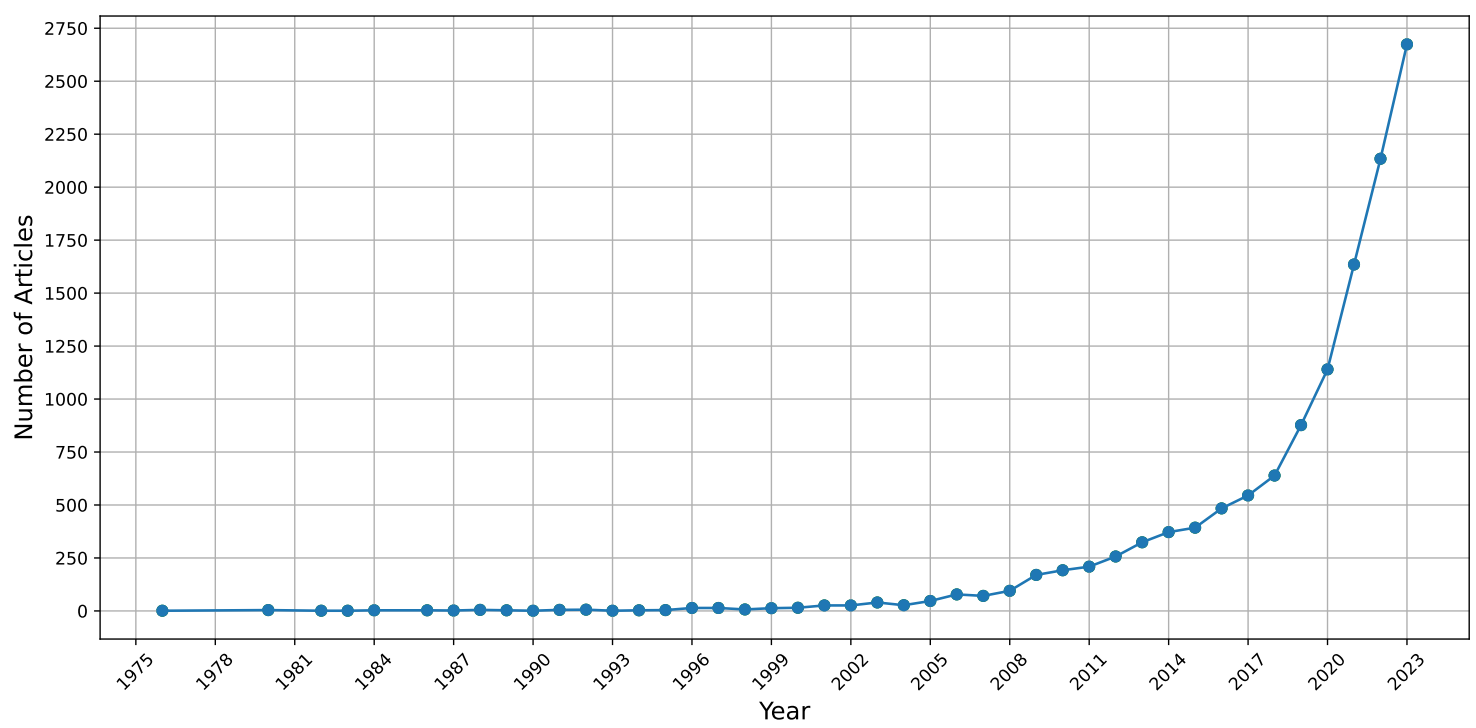

Figure 1: Annual number of articles published in machinery fault diagnosis research (Keyword “Machinery Fault"; Source: Scopus)

### 1.3. Organization

The structure of this paper follows the taxonomy developed for this literature review. Section 2 outlines the methodology and taxonomy created for this research. Section 3 discusses machinery fault data, including analysis and a summary of available datasets. In Section 4, we present data-driven approaches, covering both traditional and advanced methods, and highlight recent advancements in the field. This section briefly reviews conventional methods and explores various ML algorithms and learning paradigms applied in MFD. Sections 5, 6, and 7 address the challenges associated with machinery fault datasets and algorithms used for fault detection, offer recommendations, and outline future directions in the MFD field. The review concludes in Section 8. Additionally, the appendix (Section 9) includes a list of abbreviations with their full forms and a brief introduction to the extensive datasets used in this field.

## 2. Methodology Implemented and Taxonomy Developed

In this review, we employed a narrative overview approach to synthesize the current state of research in MFD

**Table 1**
Comparison of this article with other reviews on MFD studies.

| AR | Year | Machinery | Algorithms covered | Types of ML | Datasets Mentioned |
|---|---|---|---|---|---|
| [16] | 2018 | ○ | U/SL | MLP, AE, DBN, CNN | ○ |
| [31] | 2018 | ◑ | U/SL | kNN, NBC, SVM, ANN, AE, DBN | ○ |
| [9] | 2019 | ○ | U/SL | CNN, AE, DBN | ○ |
| [17] | 2019 | ○ | U/SL | SVM, CNN, AE | ○ |
| [8] | 2020 | ○ | U/SL, SSL | ANN, PCA, kNN, SVM, CNN, AE, DBN, RNN, GAN, TL | ◑ |
| [1] | 2020 | ○ | U/SL, SSL, RL | AE, CNN, DBN, RNN, GAN, RL, TL | ◑ |
| [14] | 2021 | ◑ | SL | SVM, kNN, ANN, DT&RF, NBC, CNN | ○ |
| [10] | 2021 | ○ | U/SL | Mentioned TML, AE, CNN, DBN, RNN | ○ |
| [13] | 2022 | ○ | SL | SVM, MLP, RF, CNN, RNN | ○ |
| [29] | 2022 | ○ | U/SL, SSL | AE | ○ |
| [24] | 2023 | ○ | U/SL, SSL | ANN, SVM, kNN, CNN, AE, GAN, RNN, DBN, TL | ◑ |
| [32] | 2023 | ● | U/SL, SSL | DBN, AE, RNN, CNN, GAN, TL | ◑ |
| Our Review | 2024 | ● | U/SL, SSL, RL | TML, AE, CNN, DBN, GAN, RNN, RL, TL, FL, AD, PINN | ● |

Notes: ○ - Fully considered ; ◑ - Partially considered; ○ - Sparsely considered. The full forms of the abbreviations used are found in Table 13.

using ML techniques. Our primary sources of literature were Google Scholar and Scopus, where we conducted comprehensive searches using a variety of relevant keywords, including “machine fault" OR “machinery fault" OR “gear fault" OR “bearing fault" OR “induction motor fault" OR “Wind turbine fault" OR “rotor fault" OR “stator fault", etc. AND "deep learning" OR "machine learning". Additionally, we performed specific searches combining the term “machinery fault" with the names of individual ML algorithms (e.g., “machinery fault using CNN") to ensure a thorough coverage of the topic. Furthermore, we utilized the snowballing technique, which involved reviewing the references of the selected articles to identify additional relevant studies. In selecting articles for this review, we prioritized recent publications, focusing on those from venues that are recognized for publishing significant research in machinery fault detection. When encountering multiple studies on similar topics, we considered citation counts as a factor to assess the impact and relevance of the work. Moreover, for datasets search, we followed the references provided in the selected articles to locate the sources. Additionally, we conducted keyword searches (e.g.,“gear fault dataset, “bearing fault dataset", etc.) to find publicly available relevant datasets for MFD.

To present a comprehensive understanding of MFD through advanced learning approaches, we constructed a taxonomy that serves as the structural foundation of this review paper. This taxonomy aims to guide readers systematically through the varied aspects of the field. Starting with an exploration of the fundamental principles of MFD and the essential background of ML, the taxonomy then directs readers through a detailed examination of diverse advanced learning techniques and their practical applications in MFD. Moreover, it extends its coverage to encompass the critical domains of data acquisition and preprocessing techniques, available datasets, and practical recommendations for creating data collection test beds. A visual representation of this taxonomy can be found in Figure 2. In this figure, topics are categorized based on the depth of coverage in this review. Topics shaded in dark green are covered in-depth, providing comprehensive explanations and insights. Topics with partial coverage are shaded in light green, indicating a moderate level of detail. Lastly, topics that are only briefly mentioned are left unshaded.

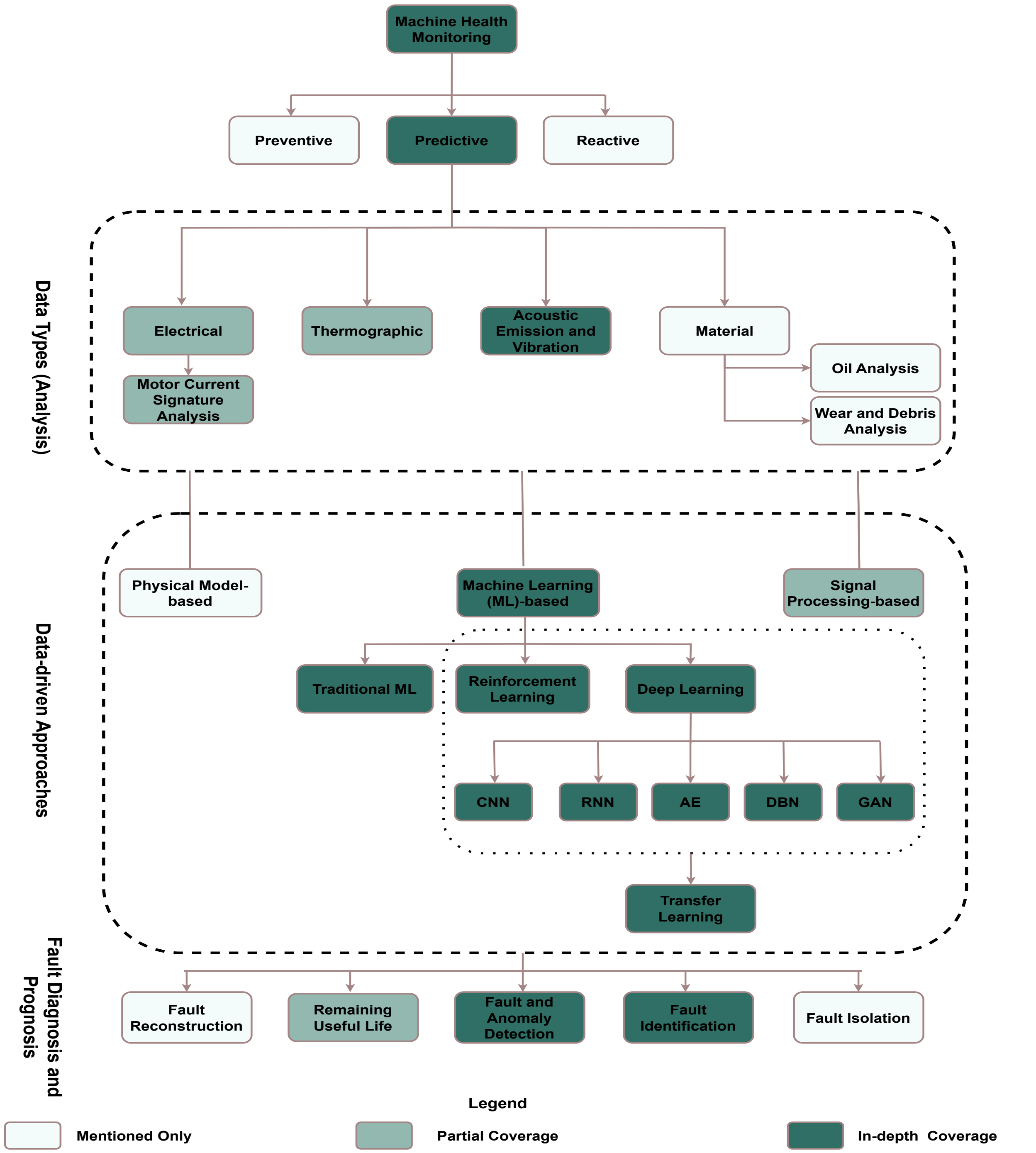

**Figure 2:** Taxonomy developed for this literature review.

## 3. Machinery Fault Data (and Analysis)

Data is a fundamental element in all data-driven approaches, and the performance of these approaches depends significantly on the amount, quality, and diversity of the data used. Selecting the appropriate target data from the initial dataset is crucial for enhancing the reliability and accuracy of the predictive model. High-quality datasets are vital in enabling these models to learn and generalize the underlying patterns and features associated with various conditions or failures. This section discusses the importance of data for effect fault analysis, the ability of advanced learning algorithms to use large amounts of data, and strategies like data augmentation that address challenges related to data scarcity. Additionally, we explore the diverse data types commonly used in this field which are collected to assess and predict the condition of machinery. A comprehensive overview of the available datasets used in these analyses is provided in the Appendix section 9. The summary of these datasets, including their key characteristics, is presented in Table 2.

**Table 2**
Summary of Datasets and Corresponding Articles Using These Datasets in MFD Research[1]

| Dataset | AR | Element | Data type | Uses | Source |
|---|---|---|---|---|---|
| Airbus | [33] | Helicopter | Vibration | Classification, TFD | AirBus |
| AI4I | [34, 35] | Synthetic | Multivariate Time-series | Classification | UCI |
| C-MAPSS | [36–39] | Turbofan engine | 21 sensor data | Prediction, TFD | C-MAPSS |
| CWRU | [40–98] | Bearing | Vibration | Classification, TFD | CWRU |
| DIRG | [99, 100] | Bearing | Vibration | Classification, TFD | Article |
| EDGFD | [101] | Gear | Vibration | Classification | Google Drive |
| FEMTO | [102–105] | Bearing | Temperature & Vibration | Prediction, TFD | GitHub |
| Gearbox Fault Diagnosis Data | [81] | Gear | Vibration | Classification | Kaggle |
| HUMS | [106, 107] | Helicopter Gear | Vibration | Early FDet, Crack Propagation Trend, AD | HUMS2023 & HUMS2025 |
| IMS | [42, 52, 68, 71, 98, 104, 108–111] | Bearing | Vibration | Classification, Prediction | IMS |
| IMTI | [112, 113] | Induction Motor | Thermal Images | Classification | Mendely Data |
| JNU | [114, 115] | Bearing | Vibration | Classification, TFD | - |
| MaFaulDa | [90] | Bearing | Vibration | Classification, TFD | MaFaulDa |
| MFPT | [90, 116] | Bearing | Vibration | Classification, TFD | MFPT |
| NEU | [117, 118] | Steel Surface Defect | Images | Classification | this website |
| NREL wind Turbine | [88, 119] | Gear | Vibration & Speed | Classification | OpenEI |
| PHM2009 | [71, 83, 120–122] | Gearbox, bearing, shaft | Vibration | Classification, TFD | PHM2009 |
| PHM2010 | | Milling Cutter | Force, Vibration, Acoustic | Prediction | PHM2010 |
| PHME Datasets | - | Various Types | Various time series | Classification, Prediction, Root cause analysis, etc. | GitHub |
| PU | [48, 67, 81, 89, 100, 114, 116, 123–125] | Bearing | Current & Vibration | Classification, TFD | PU |
| Rotor Fault | [126, 127] | Rotor | Vibration | Classification | Mendeley Data |
| SEU | [46, 72, 100, 128, 129] | Bearing, Gear | Vibration | Classification and TFD | SEU |

[1] The full forms of these abbreviations can be found in Section 9.

| | | | | | |
|---|---|---|---|---|---|
| TEP | [76, 130–132] | Chemical process | Simulation | FDet, Process Control | TEP |
| THU | [40, 133, 134] | Gear | Vibration | Classification, TFD | - |
| UA-FS | [133–137] | Gear | Vibration | Classification, TFD | - |
| UO | [80, 133] | Bearing | Vibration | Classification, TFD | Mendeley Data |
| UoC | [70, 77, 78] | Gear | Vibration | Classification, TFD | UoC |
| UORED-VAFCLS | [138] | Bearing | Vibration, Acoustic | Classification | Mendeley Data |
| Wind Turbine SCADA | [139–141, 141–144] | Wind Turbine | Speed, Power, etc. | Classification, TFD | Zenedo, Kaggle & GitHub |
| XJTU-SY | [78, 88, 145, 146] | Bearing | | Prediction | XJTU-SY |
| MFS and In-Lab Dataset | [53–55, 59–65, 68, 74, 79, 85, 87, 109, 120, 147–159] | Various elements | Various types | Classification, Prediction, TFD, etc | - |

### 3.1. Data Types and Analyses

Predictive maintenance analysis for MHM relies on various types of data collected from the machine and its equipment to predict potential failures and optimize maintenance. The primary data types used in PdM are vibration and acoustic emission data, material data (wear & debris, oil samples), electrical data (current, phase), image data (thermal images), and other types like temperature, speed, torque, and so on. Most of these data types are time-series data, where data collection, analysis, and interpretation follow a similar pathway. Firstly, **accurate sensor placement** is critical. Sensors should be mounted close to the source to capture relevant information effectively. Additionally, secure mounting is essential to minimize interference or noise that could compromise data quality. Then, **data collection and acquisition** is facilitated by various acquisition devices—ranging from standalone to smartphone-based systems—chosen based on criteria like the machinery's complexity, the need for portability, measurement accuracy, and budget, ensuring that the selected device has an adequate sampling rate, resolution, and dynamic range is critical for precise data capture across all types of time-series data. The next step is **data analysis**, where the collected data is analyzed using various techniques to identify potential machinery faults. Common methods include time-domain analysis, frequency-domain analysis, and time-frequency analysis. Machine learning and deep learning techniques, as well as physics-based approaches like finite element analysis and rotor dynamics, are also employed. Statistical methods, such as calculating standard deviation, mean, kurtosis, and skewness, are used alongside data-driven approaches to enhance fault detection and diagnostics. The **ultimate goal** is to utilize these analyses to estimate the RUL of machinery, accurately detect and classify faults, and localize or isolate them. Figure 3 represents this process.

The vibration data type is the mostly acquired and analyzed, accounting for nearly three-quarters of all PdM methods due to its effectiveness in detecting a wide range of machinery faults [160]. Other analysis methods include wear debris analysis, oil analysis, Acoustic Emission (AcE) analysis, thermography (or thermal imaging), and motor current signature analysis.

A. ***Vibration Analysis:*** Vibration analysis is a non-intrusive method sensitive to early fault stages [160]. Machines emit vibration signals that vary with condition changes, providing diagnostic information. Sensors like accelerometers capture these signals, which are then analyzed to detect and diagnose faults such as unbalance, misalignment, and bearing issues, and to predict RUL. Accurate sensor placement, data collection, and analysis are key steps in vibration analysis [161].
B. ***Wear Debris Analysis:*** This technique examines particles generated by machinery wear to identify faults. Analyzing the chemical makeup, size, color, and shape of wear particles identifies wear mechanisms like rubbing, cutting, and fatigue [149]. Collection methods include magnetic plugs, filters, and centrifugation, with analysis techniques like ferrography, spectroscopy, and particle counting. Effective in detecting early-stage faults in gears and bearings, wear debris analysis requires significant expertise, specialized equipment, and an understanding of machinery operation.

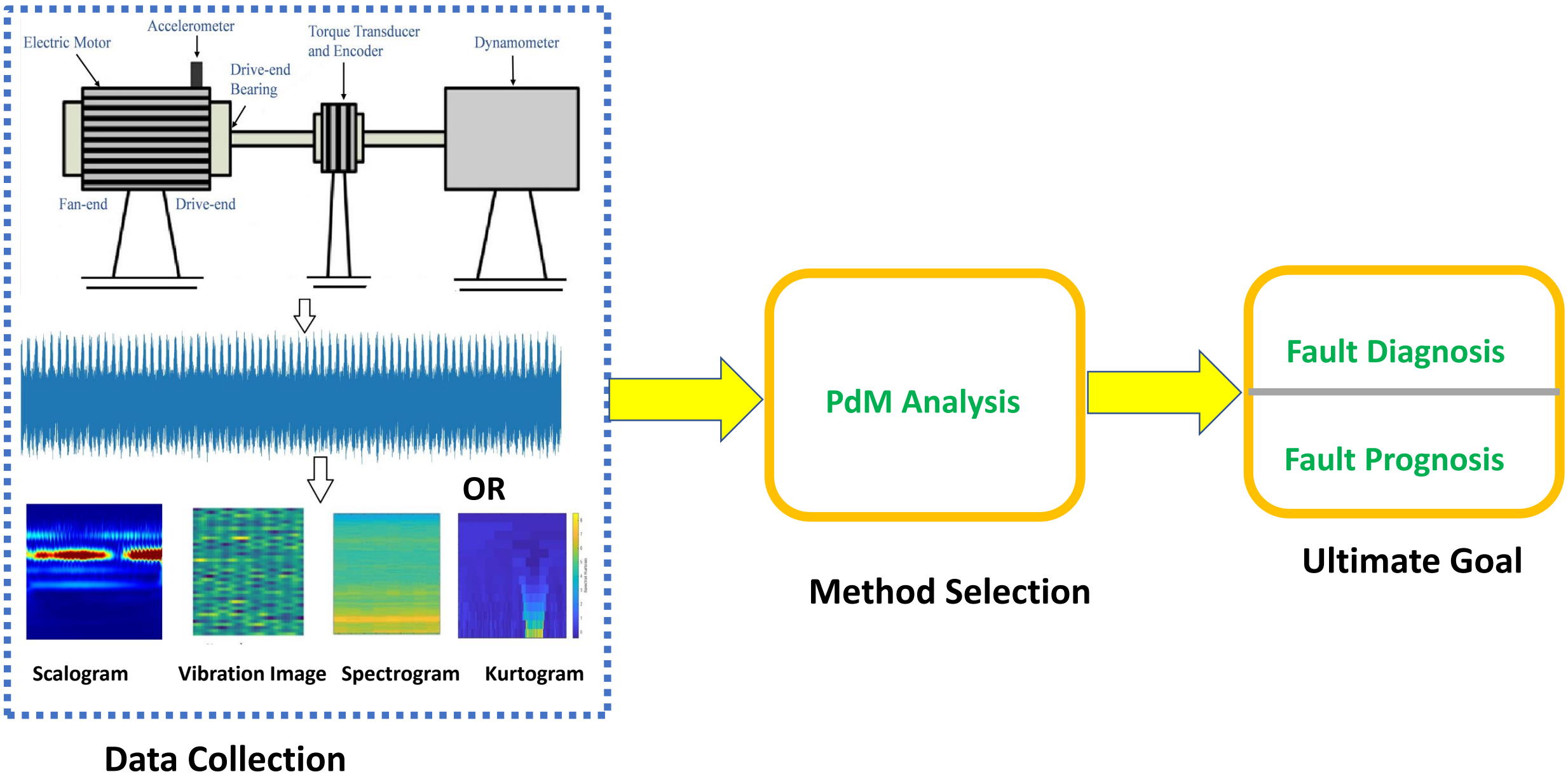

**Figure 3:** General framework of MFD using ML-based algorithms

C. ***Oil Analysis:*** Oil analysis detects signs of wear or contamination in machinery. Regular samples are tested for properties like viscosity, acidity, particle count, and elemental composition, identifying potential faults. Common in engines, gearboxes, and hydraulic systems, oil analysis, combined with other methods like vibration analysis, provides a comprehensive understanding of machinery health [162].
D. ***Acoustic Emission Analysis:*** AcE analysis detects high-frequency sound waves generated by defects or wear. AcE sensors attached to machinery detect signals processed to extract information. Common in rotating machinery and structures, AcE analysis identifies defects through sound wave patterns [163].
E. ***Thermography:*** Thermography uses infrared cameras to capture temperature distribution, indicating potential faults. Infrared cameras detect emitted radiation, creating images representing temperature variations. Analyzing these images identifies hot spots or anomalies in electrical systems, motors, gearboxes, bearings, and insulation materials [164].
F. ***Motor Current Signature Analysis:*** MCSA monitors electric motors by analyzing the current signal in motor windings, detecting faults like bearing, rotor, and stator issues. The signal spectrum analysis identifies operating condition anomalies, helping diagnose faults and optimize maintenance [165].

## 3.2. Data Augmentation

Acquiring comprehensive datasets for MFD is a challenging task due to the rarity of failures and high data collection costs, which often result in limited and biased datasets [166]. To address this issue, various techniques, such as data augmentation and transfer learning, are employed to enhance dataset diversity and improve model performance. Data augmentation involves transforming existing data in the time or frequency domains to generate new samples, effectively preventing overfitting and improving generalization [167]. On the other hand, Transfer Learning (TL), which is explained in Section 4.4, uses related domain knowledge. Moreover, Collaboration and data sharing among researchers are also encouraged to promote the development of robust MFD models [47]. These approaches help to overcome the issues of data scarcity and imbalance that can affect the accuracy of data-driven fault classification methods. A detailed list of studies that have employed these techniques can be found in the table table 3.

**Table 3**
Data Augmentation techniques used in MFD

| Augmentation Techniques | Article References |
|---|---|
| Oversampling | SMOTE [168], SCOTE [169], K-means SMOTE [170], SI-SMOTE [171], imputation-based [172], cluster-majority weighted minority-based [173], Overlapping [174], clipping [175], Gaussian Noise [19, 176] |
| Data Transformation | Cut-flip [177], time warping [178, 179], permutation [179] |
| Generative | cGAN [180], GFMGAN [181], WCGAN [182], DCGAN [183], Wasserstein GAN [184], VAE [185], cVAE-GAN [186], VMR [187], cVAE [188] |

## 4. Data-driven Approaches

In MFD, the PdM techniques can be broadly classified into Physical Model (PM)-based [189], Expert System (ES)-based, Signal-Processing-based [190], ML-based [7], and their hybrids [23], according to their developmental progression. However, for the purpose of this literature review, we have organized them into two distinct categories: (i) Traditional Data-driven Approaches, encompassing PM-based, ES-based, SP-based, and TML-based, and (ii) Advanced Data-driven methods, which include DL, RL, TL & Domain Adaptation (DA) and hybrid approaches.

### 4.1. Traditional Data-driven Approaches

A. **Physical Model-based Approach:** PM-based methods rely on the system's mathematical or analytical model. It is like knowing the blueprint of the machine. The task involves identifying defects in the processes, actuators, and sensors by leveraging the dependencies among various measurable signals [191]. PM-based methods are inefficient and inflexible for practical use because (i) they require a thorough understanding of machine mechanism, (ii) it is challenging to build accurate physical systems for modern complex mechanical devices operating in noisy environments, and (iii) they are unable to update data in real-time.

B. **Expert System-based Approach:** Similarly, MFD through expert system-based approaches relies on the automation of the diagnostic process using expert-level knowledge. These systems employ various reasoning methods, including rule-based reasoning, fuzzy logic-based reasoning, neural network-based reasoning, and case-based reasoning [7]. Despite their advantages, expert systems face limitations, such as their heavy reliance on expert knowledge, which is difficult to acquire and express, and their lack of self-learning capabilities, which restricts their ability to expand and update the knowledge base.

C. **Signal-processing-based Approach:** Furthermore, the SP-based methods aim to extract the relevant information from the collected signals to identify the potential faults in a machine. These techniques help emphasize the machine's fault status characteristics by employing advanced signal filtering and denoising methods [192]. The commonly used SP methods in MFD are Fast Fourier Transforms (FFT) [193], Wavelet Transforms (WT) [194], Wavelet Packet Transform (WPT) [79] Empirical Mode Decomposition (EMD) [195], Hilbert-Huang Transform (HHT) [196], cepstrum analysis [197], envelope analysis [198], variational mode decomposition [199], and so on. With the use of these techniques, the appropriate level of accuracy can be achieved; however, (i) these methods struggle with complex, non-linear, or non-stationary signal sources, (ii) are time-consuming, (iii) have limited predictive capability for prospective faults, and (iv) usually require a mathematical basis and deep technical knowledge for feature extraction and understanding the significance of different frequency components.

D. **Traditional Machine Learning-based Approach:** TML algorithms have been widely used in MFD. Initially, in the early 2000s, these algorithms were primarily employed for fault-state classification or identification processes. Over time, their application expanded to include integration with the SP methods and, eventually, their incorporation into deeper networks as ensemble networks. The commonly used TML algorithms for MFD are Artificial Neural Network (ANN), Support Vector Machine (SVM), Decision Tree and Random Forest (DT & RF), K-Nearest Neighbor (KNN), Naive Bayes Classifier (NBC), Hidden Markov Model (HMM), K-means clustering, C-means clustering, Principal Component Analysis [200], regression analysis [201], and so on. Table 4 presents some of the works employing TML in the field of MFD and also presents the general advantages and disadvantages of these algorithms.

**Table 4**
Summary of applications of TML-based algorithms used in MFD, and their respective advantages and disadvantages [2]

| AR | M | Advantages | Disadvantages | Model |
|---|---|---|---|---|
| [202–206] | B | High self learning capability | Increased complexity with more data | |
| [207–209] | G | Recognizes multiple machine states | Potential overfitting | ANN |
| [210–213] | IM | | Low interpretability | |
| [214, 214, 215] | B | Better interpretability compared to ANNs | Sensitive to kernel parameters | |
| [216–218] | G | High diagnosis accuracy | Complicated for multi-class tasks | SVM |
| [219, 220] | IM | | Inefficient with massive data | |
| [221, 222] | B | Easy implementation | High computation cost for large datasets | |
| [216, 223] | G | Intuitive and simple to understand | Difficult parameter (k) determination | KNN |
| [219, 224] | IM | | Reduced accuracy with imbalanced data | |
| [225–227] | B | Handles missing data | Prone to overfitting | |
| [228, 229] | G | Naturally interpretable | Often constructed based on expert knowledge | DT & RF |
| [230, 231] | IM | | | |
| [232, 233] | B | Simple and fast; Scalability | Assumes feature independence | |
| [228, 234] | G | Low training complexity | Poor performance with highly correlated features | NBC |
| [230, 235] | IM | | | |
| [236, 237] | B | Captures temporal dependencies | Computationally intensive | |
| [238, 239] | G | Can handle noisy data | Sensitive to initial parameters | HMM |
| [240, 241] | IM | | | |
| [242–244] | B | Summarizes data, reduces dimensions | May not scale well with large datasets | |
| [245, 246] | G | Understands natural groupings | Some algorithms are computationally expensive | Clustering Methods |
| [247, 248] | IM | | | |

## 4.2. Deep Learning-based Data-driven Approaches

While TML-based techniques show promise in MFD, they struggle with early fault detection in complex industrial environments. The dynamic, non-linear, and multi-modal nature of industrial processes complicates data analysis. TML methods require manual feature extraction and selection, which is challenging for large datasets and separates feature mining from decision-making, leading to inefficiencies [249]. As machine complexity and data dimensions grow, these limitations hamper the effectiveness of classical models. Additionally, traditional fault detection methods, dependent on human expertise, are time-consuming, error-prone, and inadequate for modern industrial systems [27]. The presence of noise and non-stationary machinery signals further complicates fault detection. Consequently, DL has gained significant interest in improving fault detection and diagnosis. Emerging as a powerful approach in the mid-2010s [250], DL has revolutionized MFD by learning useful representations directly from raw data, eliminating the need for manual feature engineering. This capability, combined with advancements in sensors and IoT, allows DL to process large volumes of data, improving the accuracy and efficiency of fault detection and diagnosis [29, 251].

FD in rotating machinery using advanced data-driven approaches builds on the steps outlined in Subsection 3.1 (also represented in Figure 3) The steps include: (a) collecting sensor data that can indicate the health of the equipment, (b) extracting features from the collected data using various models and algorithms, and, (c) identifying and classifying the fault state of the equipment based on the extracted fault-sensing features [32]. In these approaches, vibration, acoustic emission, temperature, and current data are commonly used, with specific methods applied to extract features from different domains (time, frequency, time-frequency). Feature selection methods then identify the most relevant features, which are used to train machine learning models for fault classification, anomaly detection, and RUL prediction. This section explores the principles, techniques, and applications of different types of DL algorithms used in MFD.

[2] AR: Article references; B: Bearing; G: Gear; IM: Induction Motor; M: Machinery type used

#### *4.2.1. Convolutional Neural Network-based FD*

CNNs are widely used for machine fault classification. In fact, MFD is one of the earliest and most extensively explored fields in CNN-based FD (CNNFD), directly inspired by the image classification principle [30]. Both 2-D and 1-D CNNs are applied in this context. 1-D CNNs focus on time series data, while 2-D CNNs handle multidimensional data with spatial/temporal correlations. Different variants of CNNs, like ResNet, DenseNet, VGGNet, capsule networks (CapsNet), dilated CNN, region-based CNN, and more, have also been adapted and deployed in MFD to enhance manufacturing and industrial processes. For an organized review, we have subdivided the study of CNNFD based on the structural characteristics of convolutional networks into two aspects: FD using 1-D CNN and 2-D CNN.

A. **1-D CNNFD:** Employing 1-D CNN for FD is a straightforward strategy in which the raw 1-D data can serve as a direct input to the CNN model. Vibration data, as previously mentioned, is the prevalent data type within this strategy. The study [110] employed a 1-D CNN (3 convolution and 2 multi-layer perceptron layers) for bearing fault classification of the IMS dataset. The raw vibration data undergoes several essential preprocessing steps. It is decimated by a factor of 8 to manage complexity, providing a bandwidth of 12.5 KHz. Low-pass filtering eliminates high-frequency components, and data is normalized for consistent scaling. The study achieved 97.1% of accuracy. Similarly, Zhang et al. in [75] employed 1-D CNN for bearing fault diagnosis using CWRU bearing dataset under a noisy and variable working environment. Utilization of dropout and small-batch training techniques, and avoidance of complex prepossessing makes this work unique. Slicing the training samples with overlapping techniques is utilized as a data augmentation method. An impressive classification accuracy of 99.77% was achieved, particularly under a signal-to-noise ratio of 10. Additionally, the t-SNE method was employed to visualize and comprehend the classification results. Moreover, in the study [76], a zero-shot FD method based on semantic space embedding was implemented. This approach utilized a 1-D CNN with two convolution layers and a fully connected layer for extracting fault features from raw data from CWRU and TEP datasets. To enable zero-shot learning, human-defined fault label embeddings were created as a fault attribute matrix with seven fine-grained fault attributes for each bearing fault. Feature embeddings were matched with fault attributes using cosine distance. Moreover, the study [111] utilized the IMS bearing vibration dataset to enhance electric motor bearing fault detection via a multichannel 1-D CNN classifier processing time-domain vibration data. Z-score standardization and linear scaling were applied to preprocess the data. The two-channel classifier achieved 100% accuracy by using both x and y-axis data, and a two-channel/two-level classifier distinguished early and advanced fault levels with 84.64% average accuracy. Similarly, in [81], a Wide-Kernel CNN (WK-CNN) was utilized to process raw time series data from three industrial machinery fault datasets: CWRU bearing, gearbox fault diagnosis, and Paderborn bearing datasets. The methodology focused on varying architectural hyperparameters (like kernel size, stride, and filters) across 38,880 model iterations, analyzing their impact on model performance. Data preprocessing involved segmenting, batch processing, and splitting into training and test sets. The results, measured primarily through accuracy scores, revealed significant performance variations across datasets and hyperparameter configurations, with some models achieving near-perfect accuracy. Additionally, ML/DL algorithms were employed to further understand the non-linear relationship between specific hyperparameters and the WK-CNN's performance, underscoring the nuanced influence of the number of trainable parameters on different datasets.

Apart from vibration data, other mechanical fault signals, like AcE, current, temperature and so on, are also used as an input for 1-D CNNFD. In [252], authors presented a blade damage identification method for centrifugal fans. They used a multi-level fusion algorithm, combining vibroacoustic signals–collected from acoustic pressure sensors and accelerometers under different fan speeds and noise levels–with a 1-D CNN network. Acoustic and vibration signals were fused at the data level using adaptive weighted fusion, followed by feature extraction with a 1-D CNN network. The extracted features were then combined through a fully connected layer. Similarly, the work [139] focused on predicting wind turbine blade icing faults by combining ReliefF feature extraction with a 1-D CNN stacked bi-directional gated recurrent unit model. Using SCADA data from China's 2017 industrial big data competition, initial data preprocessing involved reducing feature dimensions from 20 to 15 using ReliefF and reconstructing the data with a sliding time window. A weighted accuracy metric addressed data imbalance, and 5-fold cross-validation was used. Results showed a significant 43.08% increase in weighted accuracy compared to traditional models. The study [253] aimed to diagnose faults in permanent magnet synchronous motors by analyzing stator current signals. Stator current data from various fault scenarios and speeds were collected, resulting in a 2000-point dataset, which was split into 1000 for training and 1000 for testing after normalization.

Two key methods, the 1-D CNN and the WPT, were employed for feature extraction. The 1-D CNN outperformed other methods with a diagnostic accuracy of 98.8%.

With the increasing concept of explainable AI (XAI), the need for transparent and understandable models in complex mechanical systems is more critical than ever. Addressing this, the study [100] utilized a Multi-Wavelet Kernel (MWK) CCNN to analyze 1-D vibration signals from gearboxes. It employed DDS datasets provided by Southeaset University, China and Wind Turbine Gearbox Dataset. Integrating a CWT with a traditional CNN, this study introduced a MWK convolutoin layer and a kernel weight recalibration module, and employed heatmaps for visualizing learned feature maps, thereby improving the interpretability of impulse detection in gearbox fault diagnosis.. Preprocessing of the data involved signal cutting using a sliding window of size 1024 and standard normalization techniques. The MWK-CNN achieved over 98% classification accuracy in gearbox fault diagnosis under various conditions.

B. **2-D CNNFD:** Initially, CNN architectures applied for MFD imitated the 2-D structure used in image processing. Since mechanical signals are mainly 1-D time series, the main approach was to change these 1-D signals into 2-D. To do this, several SP methods were implemented, which are described briefly in the following paragraph. Many researchers convert 1-D time series data into 2-D image formats using data matrix transformations, referred to as "vibration images" [41]. Here, the raw 1-D data is transformed to 2-D matrices (image-structure). For example, if we have a 1-D signal $X = [x_1, x_2, \dots, x_m]$ and wish to convert it into a 2-D matrix of dimensions $a \times b$ (where $a \times b = m$), the transformation can be applied as $Y_{i,j} = x_{(i-1)\cdot b+j}$ for $1 \leq i \leq a$ and $1 \leq j \leq b$. This process effectively restructures the time series data into a format similar to an image, allowing for the application of image processing techniques to analyze the signal patterns. Besides 'vibration images', thermal images [254] have also been used as input for CNNs. In MFD,thermal imaging is key for spotting temperature anomalies that indicate faults, enabling non-invasive, continuous monitoring and diagnosis, often used alongside other techniques like vibration analysis to prevent major machine issues. Additionally, kurtograms [255], Scalograms [44], and Spectrograms [41] are important SP techniques in MFD, which analyze machine signal and noise. Kurtograms focus on detecting transient faults by assessing signal kurtosis across frequency bands, highlighting sudden signal changes. Scalograms derived from the continuous wavelet transform (CWT) provide a time-frequency view of the signal that efficiently highlights frequency changes over time. Spectrograms generated using the short-time Fourier transform (STFT) provide a time-frequency analysis ideal for tracking changes in machine behaviour. These techniques facilitate the detection of machine faults and potential defects by processing raw vibration or acoustic data through mathematical algorithms and converting the signal into a visual form. The data types used in 2-D CNN are represented in Table 5.

**Table 5**
2-D CNN applications in MFD according to types of inputs

| Transformation Methods | Article References |
|---|---|
| Data Matrix | [236, 256–265] |
| Vibration Image | [41, 266–273] |
| Thermal Image | [254, 274–277] |
| Time/Frequency Transformation | [82, 166, 255, 278–284] |
| Short-time Fourier Transform | [285–290] |
| Wavelet Transform | [44, 291–300] |

As an application of data matrix transformation method, Neupane et. al [41] used this technique to transform raw data to 2-D matrices and then employed a simple 2-D CNN model for the bearing fault detection of CWRU dataset. Similarly, in their innovative study, Jiao et al. [301] proposed a deep coupled dense convolutional network (CDCN) for mechanical fault diagnosis. Using a 1-D convolutional structure and dense connections, the CDCN effectively extracts features from raw, nonstationary mechanical signals. Unlike conventional data-splicing techniques, the model integrates multisensor data as parallel inputs. These sensors capture both transverse and torsional vibrations, enabling a double-level information fusion approach for more accurate fault identification. The CDCN model, achieving 99.39% accuracy, demonstrated superior recognition accuracy, convergence speed, and classification accuracy compared to traditional CNNs, single-sensor data methods, and data-splicing-based fusion techniques. All the experiments were carried out with an Intel Core i7 CPU and GEFORCE GTX 1060

GPU 10 times to reduce randomness. The authors simulated and tested nine health conditions of a planetary gearbox on a test rig, running experiments at 20Hz driving speeds with a 2 N·m load, and collected data at a sampling frequency of 5kHz, resulting in a total dataset of 5400 samples for the nine conditions; they then add Gaussian white noise to simulate a real, harsh industrial environment. Even though the model could potentially increase computational costs and introduce discrepancies between training and testing data distributions, its promising results and planned improvements indicate that this approach has contributed substantially to the progress in the field of intelligent fault diagnosis.

Alternatively, employing the statistical data from either the time or frequency domain as the input for a convolutional network is another method of conversion. In the study [302], a gear test rig with an accelerometer and high-speed camera was used to train a CNN (VGG16 ConvNet) with 2-D grayscale images from FFT spectrums of vibration signals. The dataset included 600 images, 500 for training and 100 for testing, from endurance tests on plastic gears. Employing transfer learning, the VGG16, initially trained on ImageNet, was retrained with these vibration data images. In the preprocessing step, vibration signals were transformed into images using an FFT spectrum peak picking method. This method involved selecting amplitude peaks across frequencies from 0 Hz to 1600 Hz in a zig-zag pattern, every 16.67 Hz, resulting in 12×16 pixel grayscale images. Each image, representing frequency amplitudes and phases, was then labeled as 'crack' or 'non-crack' based on high-speed camera observations, providing labeled data for CNN training. Remarkably, the study achieved 99% accuracy in training and 100% in testing, primarily aimed at detecting cracks in plastic gears, with a focus on retraining the model's final two layers. Moreover, the research detailed in [82] introduces a FD method for rolling bearings, integrating a deep CNN (DCNN) with an immunity algorithm. This technique uses 2-D images derived from bearing fault signals in both time and frequency domains as input for the DCNN. Utilizing data from the CWRU Data Center, the method combines DCNN for feature extraction and an immunity algorithm for adaptive learning of new faults. Preprocessing includes transforming vibration signals into 2-D images. The method achieves over 98% accuracy in fault recognition, effectively minimizing false positives and negatives, showcasing its proficiency in adaptive learning and accuracy in fault diagnosis through advanced machine learning. Moreover, the study [255] focuses on FD in rolling-element bearings using CNN, specifically a modified LeNet-5 architecture. It uniquely transforms 1-D AcE signals into 2-D kurtogram images for CNN compatibility. The dataset comprises AcE signals from bearings with various fault conditions and normal states. The preprocessing includes converting 1-D signals into 2-D kurtograms using a 1/3-binary tree approach. The 1/3-binary tree approach in the fast kurtogram algorithm is an extension that uses three additional band-pass filters to further decompose signal sequences, achieving finer frequency and resolution sampling with negligible extra computing cost. With this methodology, classification accuracies for different bearing conditions at various speeds exceeded 95%, with some conditions achieving 100% accuracy at 500 RPM. Furthermore, the study [80] introduces AntisymNet, a lightweight CNN for diagnosing faults in rotating machinery, transforming 1-D vibration signals into 2-D images for processing. Utilizing datasets like MiniImageNet, CWRU Bearing, Ottawa Bearing, and Hob datasets, the model achieves high accuracy (up to 99.70% on CWRU). AntisymNet's innovative architecture combines forward and reverse branches for efficient feature extraction and fusion, demonstrating reduced complexity and strong performance across different data ratios, underscoring its practical application in industrial fault diagnosis. Also, in the study [303], a 0.5 HP induction motor connected to a SpectraQuest MFS was analyzed using two accelerometers. The study employed a multi-head 1-D CNN, which analyzed 1-D vibration signals to diagnose motor faults, including bent shaft and bearing issues. The data, divided into 256-sample windows, was collected from four experimental runs. The CNN, equipped with Leaky ReLU and Early Stop features, achieved a 99.92% fault recognition accuracy.

Apart from fault detection, some research like [104] focused on the RUL prediction of rolling element bearing using CNN architecture based on similarity feature fusion. Using FEMTO and IMS dataset for validating the model, the methodology involved feature extraction, construction of similarity features based on Pearson correlation coefficient, selection of high-sensitivity features, construction of health index using PCA and RUL prediction using 1-D CNN. Preprocessing included wavelet denoising, moving average filtering, normalization, and outlier elimination.

Apart from these image types used, some researchers have employed thermal images as the input to the CNN. Li et al. [254] developed a fault diagnosis method using CNN for infrared thermal (IRT) images, captured from the machinery applying the IRT technique. The IRT images, acquired from the SpectraQuest machinery fault simulator, are selected from the thermal video to construct the data samples, and they are fed into the CNN model for fault detection. The number of fault classes used is 10, and SoftMax is used as the classifier. Similarly, in their work [164] on machine fault diagnosis, researchers employed a robust bearing test rig with a 220 V, 2 HP DC motor to simulate various bearing faults like inner race, outer race, ball faults, and lubrication issues. Data was collected using a uniaxial accelerometer for vibration and a thermal imaging camera for IRT. They adopted 2-D CNN, processing 1-D vibration data into 2-D images through CWT for scalograms and extracting thermal images via IRT. The data, converted to greyscale for computational efficiency, enabled the CNN to achieve 100% accuracy in fault diagnosis under constant speed conditions, with slightly reduced accuracy during speed variations.

**Table 6**
Articles employing CNN in MFD[3]

| AR | M | Dataset | Algorithm(s) | Result | Remarks |
|---|---|---|---|---|---|
| [158] | WTG | Own | Wavelet Packet Decomposition combined with a hierarchical CNN | testing accuracy of approximately 98.45% and inference time in milliseconds | Dataset:14,240 segments, 2048 data points each, 5120 Hz sampling, diverse gearbox health conditions |
| [257] | B | CWRU | Adaptive DCNN (2-D) | 92.84% without PP and 99.71% with PP | Decomposed vibration signals into 8 frequency bands, selected most sensitive for input; PSO method used for determining the main parameters of CNN |
| [84] | B, G | CWRU, Own (UESTC) | Dual convolution capsule network (1-D) | ~100% accuracy with CWRU dataset | Wide convolution kernel in first layer and narrow kernels in second; Employed k-fold cross-validation in training |
| [166] | B | Own | CNNs (2-D) | 93.16% with CNN and 87.25% with RF and SVM | Scaling accelerometer signals, windowing the data, and applying the Discrete Fourier Transform (DFT) to the signals |
| [121] | G | PHM2009 | CNN embedded with a health-adaptive time-scale representation | 99.24%, using the hyperparameters of 64 scales and 8 channels | Developed HTSR method to convert 1-D vibration signals into 2-D format; Employed multiscale 1-D convolutional filters as adaptive basis functions |
| [304] | G | Simulated helical gearbox fault dataset | Compressive sensing-based dual channel CNN (2-D) | Average accuracy of 99.39% | Utilized Modulation Signal Bispectrum for acoustic signal analysis and converted thermal images to grayscale, computing MSB magnitudes from these signals; Utilized a mobile phone for non-contact measurements of thermal images and acoustic signals |
| [256] | B | CWRU | Hierarchical Adaptive DCNN (2-D) | 99.7% testing accuracy | 1-D raw signal to 2-D image matrices manipulation |
| [267] | B | CWRU | CNN | 100% (with 30 and 60 kernel size in the first and second convolutional layer, respectively) | vibration images are created using matrix manipulation; Evaluation with varying kernels in convolution layers, different load conditions, and noise conditions |
| [85] | B | CWRU, In-lab (AT) | weight-shared capsule network (WSCN), optimized with a margin loss function and an agreement-based dynamic routing algorithm (1-D CNN) | 98.6% accuracy (CWRU) | WSCN: longer training time than traditional models; Weight-shared architecture: Fewer parameters, quicker training, better efficiency |

[3]AR: Article Reference; AT: Automobile Transmission; B: Bearing; CP: Centrifugal Pump; M: Machinery; PP: Preprocessing; PSO: Particle Swarm Optimization; R: Rotor; UESTC: University of Electronic Science and Technology of China; WTG: Wind Turbine Gearbox

| | | | | | |
|---|---|---|---|---|---|
| [86] | CP, B | In-lab, CWRU | Hierarchical symbolic analysis combined with CNN (1-D) | 98.50% accuracy in a multi-load dataset | PP: Data decomposed into frequency components through hierarchical analysis; Comparative analysis includes STFT-CNN and CWT-CNN methods |
| [305] | WTG | Simulation | Multiscale CNN (1-D) | 96.76% to 99.59% for different scales | Coarse-graining of raw vibration signals to represent them at multiple time scales |
| [87] | B, R | CWRU, In-lab | Hierarchical training using 2-D CNNs with magnet-loss pretraining for parameter initialization and feature distribution, followed by global fine-tuning | 96.56% (CWRU), 94.28% (own) | PP: Noise filtering, abnormal data exclusion, frequency domain conversion, and computation of 15 statistical features from vibration data |

Employing 2-D CNN with Explainable artificial intelligence (XAI) principles, the study [100] introduces a multilayer wavelet attention CNN (MWA-CNN) for MFD, merging CNNs with wavelet transform techniques. It employs a discrete wavelet transform layer and frequency attention mechanism to enhance noise robustness and interpretability. The methodology alternates between DWT and convolutional layers for signal decomposition and feature learning. Data from high-speed aeronautical and motor bearings (DIRG and PU bearing datasets) were used, with Z-score normalization for preprocessing and sliding segmentation for data augmentation. Notably, MWA-CNN achieved high diagnostic accuracies (98.75% at 4 dB SNR and 87.61% at -4 dB SNR). This approach improves interpretability, aligning with XAI principles, by enabling the network to focus on relevant feature information, thus enhancing decision-making transparency.

Some of the other works using CNN have been summarized in Table 6

### 4.2.2. Recurrent Neural Network-based FD

RNNs are designed to handle time-series data by processing inputs sequentially and using feedback loops to remember past states, making them ideal for analyzing temporal characteristics of machinery signals like vibration and temperature. This ability enables effective fault detection and prediction of RUL. However, RNNs often struggle with vanishing and exploding gradients, which limits their capacity to learn from long sequences [1]. This has led to the development of advanced versions such as Long Short-Term Memory (LSTM) units and Gated Recurrent Units (GRUs), which incorporate gated mechanisms to improve learning from long-term dependencies, crucial for reliable fault diagnosis in machinery.

- **LSTMs and GRUs:** LSTMs are designed to solve the vanishing gradient problem of RNN. They employ a more complex cell structure with a memory cell, input gate, output gate, and forget gate, allowing them to learn and retain long-term dependencies in the input sequences. GRUs (Gated Recurrent Units) use a simpler architecture than LSTMs, combining the input and forget gates into a single update gate and merging the cell state with the hidden state. This makes GRUs computationally more efficient while maintaining similar performance to LSTMs [306].

- **BiRNN:** BiRNN (Bidirectional RNN) combines two RNN layers with opposite information flows, enhancing sensitivity to time sequences. It considers past and future data, making it valuable for dynamic mechanical fault diagnosis, where capturing information across time is vital for identifying fault types and severity. This approach significantly improves fault identification [39].

- **Hybrid Methods:** Moreover, hybrid methods like RNN encoder-decoder (handles variable-length sequences, encoding inputs into embeddings and decoding them to outputs of varying lengths) [36, 37, 103, 307], Convolution RNN (CRNN) models (CNNs extract spatial features and RNNs analyze temporal dependencies) [130, 308–310], HMM-RNN [306], and so on are also employed in MFD.

**Table 7**
Articles employing RNN in MFD.[4]

| AR | M | Dataset | Algorithm(s) | Result | Remarks |
|---|---|---|---|---|---|
| [130] | CPF | TEP | CNN (feature learning)-LSTM (time delay capture) | 98.88% accuracy | transformed multi-variate time series to 2-D via sliding window |
| [310] | 3-phase IM | Simulation | CRNN | Improved total harmonic distortion | PP: Wavelet transform |
| [83] | B, G | CWRU, PHM2009 | Bi-LSTM for feature denoising and fusion, and a Capsule Network for pattern recognition and fault diagnosis | 98.95% diagnostic accuracy with CWRU dataset | PP: normalization and segmentation |
| [131] | CP | TEP | Bi-GRU for dynamic data processing and cost-sensitive active learning for class imbalance and unlabeled data exploration | 96.6% average accuracy in TEP dataset | PP: Batch normalization, L1-norm, and dropout |
| [311] | B | In-Lab, CWRU | Low-delay lightweight RNN, utilizing LSTM and Just Another NETwork (JANET) cells to reduce RNN"s computational cost | Yielded $\tilde{1}$00% accuracy with mechanical vibration signals and $\tilde{9}$5% accuracy using MCSA | PP: signal segmentation |
| [88] | B (WT) | CWRU, XJTU-SY, NREL | Multi-scale CNN and Bi-LSTM | mean F1 score of 97.12% | feature extraction using multi-scale coarse-grained algorithm, 1-D CNN for learning, BiLSTM for long-term data dependencies, classification with fully connected layer and softmax |
| [141] | WT-Bl | SCADA | Implemented GSDE model combining CNN with RNN, LSTM, GRU; trained on Chi-square test-based training sets | GSDE model achieved highest overall average rank | Used cost-sensitive learning for DNNs and applied GMDH with cost-sensitive criterion for ensemble predictions |
| [312] | G (DP) | In-Lab | RNN with simple recurrent units (SRU) | Recognition rate for various faults exceeded 98% | Utilization of SRU and stacked DAE for dimensionality reduction, denoising, and effective handling of complex datasets |

In work on fault diagnosis for reciprocating compressors by [313], researchers optimized LSTM models using Bayesian optimization. The method involved preprocessing vibration signal data from a single sensor to maintain temporal detail while reducing dimensionality. This process, coupled with artificial data augmentation, enhanced the LSTMs' ability to detect 17 distinct fault conditions, outperforming traditional and advanced deep learning techniques. The models' diagnostic accuracy was evaluated using confusion matrices, focusing on metrics like accuracy, precision, recall, and F-score. The best-performing LSTM model achieved a notable 93% average accuracy. Similarly, a deep LSTM (DLSTM) network-based method for predicting the RuL of NASA turbofan engines, leveraging datasets FD001 and FD003 from the C-MAPSS platform, was developed in the research [38]. The DLSTM model employed multi-sensor data fusion, grid search optimization, and the Adam optimization algorithm, and prevented overfitting with a dropout method. Preprocessing includes noise reduction via exponential smoothing and feature selection based on signal correlation and monotonicity, which led to superior performance indicators such as the lowest Score, R-value, and RUL error range. Again, the study [306] introduces a model for slewing bearing life prediction using an improved GRU network optimized by Moth Flame Optimization, paired with an HMM for early degradation detection. The methodology includes signal preprocessing using Hilbert transform with Robust Local Mean Decomposition and feature extraction in time and frequency domains. This model is verified against standard machine learning methods, demonstrating superior performance in accuracy and robustness for trend prediction and residual life prognosis of excavator slewing bearings. The mean accuracy of 92% was achieved using this model. Furthermore, the study [39]

[4]CPF: Chemical Process Faults; DP: Drilling Pump; GMDH: Group Method of Data Handling; GSDE: GMDH-based Selective Deep Ensemble; IM: Induction Motor; WT: Wind Turbine, WT-Bl: Wind Turbine Blade

employed a multiple-time window-based CNN bidirectional-LSTM to account for inconsistent lengths of condition monitoring data in the industry and predict the RUL of turbofan engines using the NASA C-MAPSS dataset. The methodology employed data preprocessing that involved feature selection, normalization, and label rectification with a piecewise linear function. Varying time window sizes were used to capture diverse temporal dependencies, enhancing the model's performance over fixed-window methods. A weighted average approach was utilized to aggregate the outcomes from multiple base models, optimizing the ensemble framework's performance.

In the study [314], a DL framework using a BiConvLSTM network, combining the strengths of CNN and LSTM networks was develop for diagnosing faults in planetary gearboxes. Data for the study was collected from a planetary gearbox test rig built by University of New South Wales [315], involving 252 tests across various fault types and conditions. The data, segmented into 2-D matrices, was processed using a bidirectional-convolutional LSTM. This approach outperformed baseline methods like ConvLSTM, CNN-BiLSTM, and others, achieving an overall classification accuracy of 84.72% with 100% accuracy in identifying fault types and locations.

### *4.2.3. Autoencoder-based FD*

Autoencoders are a type of neural network used for unsupervised learning and play a vital role in MFD due to their ability to extract complex features from data and detection of anomalies due to their encoder-decoder structure. AE-based methods are easy to implement and train. They also serve as a nonlinear dimensionality reduction tool by having fewer hidden nodes than input nodes, outperforming kernel PCA.

AEs are crucial for unsupervised learning in MFD due to their encoder-decoder structure, which is effective in feature extraction and AD. As a nonlinear dimensionality reduction tool, AE outperforms methods like kernel PCA, particularly when configured with fewer hidden than input nodes. They highlight deviations from normal conditions, indicating faults and aiding in the classification of fault types and component diagnostics. This is enhanced by variants such as sparse AE (SpAE), stacked AE (StAE), denoising AE (DAE), variational AE (VAE), and contractive AE (CAE), which handle multimodal and noisy data effectively. Furthermore, incorporating advanced neural modules like CNNs for image data and RNNs for time series allows for dynamic representation learning, making AE a robust choice for ensemble learning to improve the generalization of MFD models [28]. The AE variants employed in MFD are:

- **SpAE:** SpAEs enforce sparsity constraints on the activation of hidden layers, compelling the model to learn a compact and robust representation of the input data. In Sparse Autoencoders, the regularization term is essentially the Kullback-Leibler (KL) divergence, which measures the discrepancy between the distribution of the hidden layer's activations and a predetermined target probability distribution [29]. Some of the studies using SpAE in their work are [89, 90, 124].

- **DAE:** DAEs are designed for robust feature extraction by being trained to reconstruct original, uncorrupted data from inputs deliberately corrupted with artificial noise, thus enhancing the model's resilience to disturbances. Studies [124, 316, 317] used DAE in their work.

- **VAE:** VAEs are generative models that combine an inference network with a generative network to map input data to a probabilistic latent space. They frame feature extraction as a variational Bayes inference problem, optimizing a likelihood function to effectively learn a probabilistic relationship between observed data and latent representations. This approach imposes a structured probabilistic model on the latent space, where the encoder infers distribution parameters, allowing for robust and meaningful feature generation [28]. [146, 185, 318–320] are some of the works using VAE in their research.

- **CAE:** CAEs add a penalty term to the classical reconstruction loss function, which helps the model incorporate a regularization term that promotes smooth mapping between input data and latent space. The resulting model exhibits greater robustness to small perturbations in the input data and better generalizes to novel fault patterns. [321–324] are some of the research employing CAE for machinery fault diagnosis using intelligent systems.

- **StAE:** StAEs enhance robust feature extraction by incorporating a regularization term, specifically the squared Frobenius norm of the Jacobian matrix, into the loss function. This term penalizes sensitivity to input variations, thereby encouraging the model to learn smooth mappings from input to latent space and improving its generalization to new patterns, especially in the presence of noise. Before 2018, StAE's general applications were extensively researched [28]. They have been successfully applied to MFD, including, but not limited to [316, 318, 321].

- **Modified and hybrid versions:** Researchers have proposed various AE modifications and hybrids for improved MFD. Examples include using RNNs or LSTMs with AE to capture temporal dependencies. Convolutional AEs (ConvAE) leverage Convolutional Neural Networks (CNNs) for spatial or temporal data, enhancing performance in fault detection [325–328]. Hybrid approaches incorporating clustering or classification algorithms also enhance fault detection and diagnosis systems [329–331].

**Table 8**
Articles employing AE in MFD.[5]

| AR | M | Dataset | Algorithm(s) | Result | Remarks |
|---|---|---|---|---|---|
| [133] | B,G | THU,UA-FS Gearbox, UO Bearing | Speed normalized AE, normalizing the vibration data to remove the effects of speed vibrations for fault detection | Significant improvement in FD, with highest AUC 0.9704 ± 0.0087 | PP: low-pass filtering, downsampling, segmentation, and outlier removal |
| [90] | B, G, TBl | CWRU, MFPT, MaFaulDa, UNSW TBl | Deep sparse AE with Grey Wolf Optimization | 100% accuracy in other datasets, 95% on MaFaulDa | Rprop for training |
| [124] | B, G | PU, Qianpeng Company | Denoising integrated sparse AE: feature enhancement and denoising based on fault sensitivity, data decoupling, and adaptive loss function | Over 99% test accuracy | PP: Ensemble empirical mode Decomposition; Emphasis on dealing with weak signals and background noise |
| [328] | B | In-Lab | Bayesian optimization and channel-fusion-based ConvAE | more than 99% of accuracy, precision, recall and f1 score | PP: Vibration signals segmented using a sliding window technique |
| [145] | B | XJTU-SY | Deep spatiotemporal fusion AE network, incorporating multiscale convolution, convolutional LSTM, and attention mechanism; Integration of multimode samples for unsupervised health indicator construction, introducing a quadratic function-based shape constraint | Average comprehensive score of 0.7327 on bearing dataset | PP: Wavelet threshold denoising and Sparrow Search Algorithm-Variational Mode decomposition; Input Method: Coupling of vibration signal, vibration frequency domain signal, Intrinsic Mode Function, and its frequency domain signal into a 2-D matrix, forming multimode-coupled samples |
| [146] | B | XJTU-SY, In-Lab | Deep order-wavelet convolutional VAE, integrating improved energy-order analysis and Wavelet Kernel Convolutional Block; Employed frequency-weighted energy operator, anti-symmetric real Laplace wavelet, and multi-objective gray wolf optimizer | Average identification accuracy 99% | Improved energy-order analysis for transforming time-domain vibration signals into angle-domain signals |

#### *4.2.4. Deep Belief Network-based FD*

Deep belief networks, introduced by Hinton et al. in 2006 [332], are a type of DNN constructed hierarchically by stacking multiple layers of restricted Boltzmann machines (RBMs). The first layer of the DBN comprises an RBM that models the input data, while subsequent layers consist of RBMs that capture hidden representations derived from the preceding layers. This architecture facilitates unsupervised pre-training, initializing network weights, and biases using contrastive divergence or persistent contrastive divergence. This pre-training strategy mitigates overfitting and enhances the model's ability to generalize to unfamiliar data. The primary components of RBMs include binary random variables for the visible units and hidden units.

[5] Rprop: Resilient Backpropagation; TBl: Turbine Blade

**Table 9**
Articles employing DBN in MFD.[6]

| AR | M | Dataset | Algorithm(s) | Result | Remarks |
|---|---|---|---|---|---|
| [93] | B | CWRU | Self-adaptive DBN, optimized using the Salp Swarm Algorithm | overall classification accuracy of 94.4% | Extract the time domain, frequency domain, and time-frequency domain features |
| [125] | B | PU | Mixed pooling DBN | 98.84% accuracy | Morlet wavelets for generating time-frequency images |
| [91] | B | CWRU, In-Lab | Bi-directional DBN with forward training for feature learning and reverse generation for sample synthesis; Quantum Genetic Algorithm for parameter optimization | Average accuracy percentages: 96.57%, 94.58%, 93.74% for different imbalance ratios | PP: Normalization and truncation of data; noise time-shift layer added to reduce sample similarity; DataAug: Reverse generation part of Bi-DBN synthesizes supplementary samples to address imbalanced datasets |
| [132] | CP | TEP | DBN with extended RBMs, integrating features from raw data for layer-wise feature extraction and fault classification in chemical processes | Best average accuracy of 94.31% | DataAug: Dynamic data augmentation to capture temporal correlations |
| [92] | B | CWRU, QPZZ | CWT for transforming raw data to RGB images and employing Gaussian Convolutional DBN for fault classification | Average accuracy of 99.57% | PP: Scalogram construction from vibration signals, RGB image transformation, and whitening processing of images |
| [333] | G | Simulation | DBN trained with labeled gear fault signals, Sparrow Search Algorithm for optimizing DBN parameters | Average detection accuracy of 96.18% | PP: 13 time-domain features are extracted from each sample group of raw data |
| [96] | B | CWRU, Hydraulic Pump | Hilbert envelope spectrum and DBN | Classification accuracy of 99.55% | PP: Resampling of vibration signals, application of anti-aliasing filter |
| [142] | WTB | SCADA dataset | DBN with back-propagation and layer-wise training, Exponentially weighted moving average for monitoring prediction errors; binary vectors for fault classification | AUC of 0.88 obtained with LS-SVM classifier | Parameters reduced from 33 to 12 using domain expertise; wrapper with genetic search, boosting-tree algorithm, and relief algorithms were used for final selection in generator bearing temperature prediction |

DBNs excel in extracting fault features from specific data representations, and their applications in MFD are growing. The study [94] presented a method for diagnosing bearing faults using DBN. Using 1-D input data from the CWRU bearing dataset, multiple DBNs with different hyperparameters are constructed, which are integrated using an improved ensemble method. The DBNs utilized binary and Gaussian units and were trained using Contrastive Divergence. The method, tested by cross-validation, achieved an accuracy as high as 96.95%, proving its effectiveness in fault diagnosis under challenging conditions. Moreover, in the work [334], an optimized DBN with an improved logistic sigmoidal unit for the fault diagnosis of wind turbine gearboxes is proposed. A dataset of vibration signals from an artificially faulted gearbox is used, and the Morlet wavelet transform, kurtosis index and soft thresholding are used for SP. An improved sigmoidal unit improves convergence speed and classification accuracy: when tested on the MNIST database, for the validation, and gearbox failure data, the method outperforms conventional units, achieving high accuracy up to 96.32% and fast convergence. Furthermore, Yan et al. in [95] proposed the multiscale cascaded DBN for fault detection in rotating machinery using raw vibration signals from the CWRU bearing dataset and North China Electric Power University gear vibration dataset. Employing an improved coarse-grained multiscale process

[6] AUC: Area Under the Curve; DataAug: Data Augmentation; LS-SVM: Least Square Support Vector Machine; WTB: Wind Turbine Bearing

and a three-layer DBN for feature extraction, the proposed method used data segmentation and Fourier spectrum calculations as preprocessing techniques, obtaining over 99% classification accuracy.

Another study by Tao et al. [335] focused on bearing fault diagnosis using a DBN architecture, consisting of multiple layers of RBM and backpropagation neural network (BPNN), and multisensor information fusion. The method involved acquiring multiple vibration signals from various faulty bearings, extracting time-domain characteristics from these signals, and then inputting the data into the DBN to generate a classifier for fault diagnosis. Experiments were carried out on the QPZ-II dataset, and the comparisons were made with SVM, KNN, and BPNN methods, demonstrating that the DBN-based method achieves higher identification accuracy of 97.5% for training samples and 95.5% for testing samples. Moreover, the study [336] focused on using three different architectures: deep Boltzmann machine, DBM and StAE. The method involved the pre-processing of vibration signals through four schemes covering the time, frequency, and time-frequency domains. The dataset includes seven failure patterns of rolling bearings collected from rotating machinery systems. The performance of the DNN model is evaluated in terms of accuracy, which is more than 99% in the best setting.

Apart from the classification, DBN is used in predicting RUL of rotating machinery. The study [105] employed DBN, combining with local linear embedding (LLE), and difussion process for predicting RUL of bearings of FEMTO dataset. DBN was used for feature extraction and LLE for health index (HI) determination. HI evolved based on difussion process and a probability density function of the predicted RUL was derived in terms of the first hitting time. Moreover, a study [337] proposed an unsupervised learning-based fault diagnosis model for rotating machinery, integrating the SpAE, DBN, and binary processor. The method used SpAE to encode signals in the frequency domain, which were then processed by a binary processor and fed into a DBN for fault diagnosis. The process did not require labeled training data making it entirely unsupervised. The dataset used was the CWRU bearing dataset and gear pitting dataset, from which time domain signals were transformed into normalized frequency domain signals using FFT. During preprocessing, a binary processor converted the SpAE outputs to binary data, thereby improving the RBM's efficiency within the DBN.

#### 4.2.5. Generative Neural Network-based FD

A Generative Adversarial Network (GAN), introduced by [338] in 2014, is an outstanding unsupervised generative algorithm that learns to create realistic data from a random distribution. GAN, which is acknowledged as one of the 'Top Ten Global Breakthrough Technologies' [26], consists of a generator ($G$) that produces synthetic samples from random noise and a discriminator ($D$) that distinguishes real instances from synthetic ones. Both the generator and discriminator are designed as deep neural networks. The generator maps a latent variable to the data space using its parameters. The discriminator estimates the probability that a sample is genuine or fake using its parameters. This setup forms a two-player minimax game.

GANs have been used in many research fields, including natural language processing, computer vision, and so on. They were initially used in fault diagnosis as for data augmentation, a strategy to generate additional instances with the same data distribution to solve the problem of small sample size. Then, GANs were adapted to adversarial cross-domain fault diagnosis, known as adversarial domain adaptation, which differs from data augmentation as it uses adversarial training with both target and source domain data to extract domain-invariant features. Moreover, these algorithms have also been applied to semi-supervised learning and anomaly detection [120, 339]. Various improvisations and improvements have yielded different variants of GANs, which have been developed and practiced to solve the existing limitations, like model collapse and imbalance training. The variants can be categorized into two categories according to the improvements made, and are called structure-focused improvements and loss-focused improvements [26].

I. **Structure-focused Improvements:** GANs, based on the improvements in their structure are further divided into three categories: convolution-based, condition-based, and semi-supervised GANs. To address the original GAN's feature extraction and training inefficiencies, the deep convolutional GAN (DCGAN) was developed [183]. It employs convolutional and deconvolutional layers in the discriminator and generator, improving stability and utilizing weight sharing and local connections for enhanced performance. On the other hand, condition-based GANs, including the conditional GAN (CGAN) [180], InfoGAN [340], and Auxiliary Classifier GAN (ACGAN) [341], tackle standard GANs' mode collapse issue. CGAN uses class information for guided generation, InfoGAN employs a latent code and an extra classifier to enhance input-output correlation, and ACGAN integrates an auxiliary network for classifying faults and distinguishing real from synthetic data. Furthermore, the Semi-Supervised GAN (SSGAN) efficiently uses unlabeled data in low-labeled-data scenarios.

**Table 10**
Application of GANs in MFD

| GAN as: | Article References |
|---|---|
| Data Augmentation and Balancing | [55, 58–62, 180–184, 186, 347–356] |
| Anomaly Detection | [63, 108, 128, 153, 357–361] |
| Semi-supervised adversarial learning | [64, 65, 109, 120, 154] |
| Adversarial Training for Transfer Learning | [66–68, 155] |

It features a softmax classifier in the discriminator for distinguishing real from synthetic inputs and classifying real samples, enabling semi-supervised learning [340].

II. **Loss-focused Improvements:** Loss-focused GANs were developed to stabilize training and address issues like unstable gradients and mode collapse in standard GANs. Wasserstein GAN (WGAN) [342] uses Wasserstein distance for a more stable divergence measurement between real and fake samples, though limited by weight clipping. The WGAN with Gradient Penalty [184] further refines this by adding a gradient penalty to address these limitations. The Least Squares GAN (LSGAN) [343], proposed by Mao, employs least squares loss functions to stabilize training by penalizing samples based on their proximity to the decision boundary. Additionally, the Energy-Based GAN (EBGAN) [344] and Boundary Equilibrium GAN (BEGAN) [345] innovate with autoencoder and encoder-decoder structures in their discriminators, respectively, the latter featuring an equilibrium enforcing algorithm. These variants collectively enhance the standard GAN framework by focusing on loss function and training strategy modifications to resolve specific challenges.

The application of GANs, also tabulated in 10, in MFD can be classified as:

- **Data Augmentation and Balancing:** Recent advances in GAN-based data augmentation and class balance help to solve the problem of small samples and data imbalance in fault diagnosis.For the data augmentation, the standard procedure is to collect various fault state data, train the GAN with real instances, and train a classifier by combining the generated synthetic and real data. GANs are mainly used for mechanical fault diagnosis, especially with vibration signals collected by sensors, and from limited data, fake samples are generated. These methods can be classified into 1-D time domain, 1-D frequency domain, 2-D image signals, and 1-D feature sets for generating synthetic data [26, 346].

- **Anomaly Detection:** The AD in GAN-based MFD is becoming increasingly important, especially when only normal operating data are available. The method relies on learning with normal samples to draw boundaries that distinguish between normal and faulty states. Modern approaches use adversarial learning to improve this process and move away from manually generated features. The core idea is to use GANs to generate synthetic samples and use the reconstruction loss between these samples and the original normal samples to detect anomalies. This approach has gained acceptance not only in FD but also in other areas requiring efficient and reliable AD [128, 339].

- **Semi-supervised Adversarial Learning:** In situations where labeled data is scarce, semi-supervised learning is employed to utilize unlabeled data for model training. This approach combines unlabeled data with adversarial strategies to enhance training [64, 120].

- **Adversarial Training for Transfer Learning:** Adversarial training is applied to transfer learning, called adversarial domain adaptation (ADA), which uses source domain data to enrich limited target domain data. ADA models fall into two types: adversarial discriminative models, which create domain-invariant features for fault classification, and adversarial generative models, which facilitate domain adaptation by learning data distributions or transforming data between domains [66, 67].

## 4.3. Reinforcement Learning-based FD

Reinforcement learning, a sub-field of ML, is a computational technique that focuses on training agents to make decisions in an environment by interacting with it and learning from the feedback received as rewards or penalties. RL has its roots in two primary research domains: the first being optimal control through the utilization of value functions and dynamic programming, and the second drawing inspiration from animal psychology, particularly the concept of trial-and-error search [362]. RL distinguishes itself from supervised learning in the context of connectionism [363]. In RL, the feedback signal received from the environment assesses the action's effectiveness, rather than instructing the system on how to produce the correct action [147]. The primary goal of an RL agent is to learn an optimal policy that maps states to actions, maximizing the expected cumulative reward over time. This learning process can be achieved through various algorithms, such as Q-learning, Deep Q-Networks (DQN), Proximal Policy Optimization (PPO), Actor-Critic methods, and so on [364].

In recent years, there has been a growing interest in applying RL to machinery fault detection, diagnosis, classification, and RUL prediction. RL has found applications in diverse areas, including fault diagnosis in transmission lines [365], optimizing operations in the smart grid [366], fault detection in Hydraulic press [367], industrial process control [362]. Moreover, the use of RL can be observed in both offline and online scenarios. In the offline context, researchers have explored how RL can be leveraged for manufacturing industries, as evidenced by studies [368, 369]. Offline RL often involves training a model on historical data to make predictions and decisions [370]. On the other hand, RL is also applicable in online scenarios, where it operates in real-time to identify faults and take corrective actions promptly. The utilization of RL in online settings is a dynamic and evolving area. On the other hand, RL is also applicable in online scenarios, where it operates in real-time to identify faults and take corrective actions promptly. The utilization of RL in online settings is a dynamic and evolving area. Studies [116, 371] provide insights into this aspect. The choice between offline and online RL in machinery fault detection depends on factors such as the nature of the machinery, the availability of real-time data, and the desired response time. Both approaches offer valuable contributions to the field and are continually evolving to enhance industrial processes, improve equipment reliability, reduce downtime, and ultimately advance the overall efficiency and safety of machinery operations.

The utilization of reinforcement learning in machinery fault detection, diagnosis, and classification involves training RL agents to make decisions based on the observed data, such as vibration signals, acoustic emissions, or thermal images. The RL agent can learn to identify different fault conditions by interacting with the data and receiving feedback in the form of rewards or penalties. In most of the MFD studies, fault diagnosis is approached as a classification task resembling a guessing game. The researchers created a simulation environment resembling a 'fault diagnosis game'. It presents questions with fault samples and labels to the agent, which must diagnose them. In a K-class fault diagnosis scenario, the guessing action space ranges from 0 to K-1, with 0 denoting normal and k representing the $K^{\text{th}}$ fault type. The agent receives rewards for accurate guesses and penalties for incorrect ones. Over multiple rounds of this guessing game, the agent aims to learn an optimal policy for fault identification using sensor data from monitoring equipment. A similar approach is also seen in [69], in which the researchers employed the RL method for intelligent fault diagnosis for rotating machinery. The agent, constructed using a stacked autoencoder, learns fault diagnosis through a DQN. The method combines RL and DL, enabling end-to-end fault diagnosis for machinery, in which Memory replay and rewards help the agent learn fault mappings from raw vibration signals with minimal external guidance.

Other than the classification task, the researchers proposed an automatic neural architecture search (NAS) approach using reinforcement learning in [73]. This study employed a special RNN unit called *Nascell* within a generator network and a controller layer composed of two stacked Nascell units. The controller outputs parameters for constructing CNN architectures, including convolutional kernel sizes, kernel numbers, and pooling sizes for each layer. The process involves iteratively generating network configurations, training child models, and maximizing model accuracy as the reward through policy gradient updates. This iterative approach continues until the search converges to an optimal architecture, occasionally injecting randomness to prevent getting stuck in local optima. In short, the controller acts as an agent, shaping the CNN architecture, and reinforcement learning is used to navigate the architectural search space toward higher-performing models. Moreover, In a study [157], multi-label transfer reinforcement learning (ML-TRL), for compound fault diagnosis in bearings, is employed. It combines deep RL (DRL) and TL to enhance fault feature extraction and improve accuracy in recognizing compound faults. ML-TRL outperforms traditional methods, and its use of transfer learning involves pre-training convolutional layers, reducing the complexity of DRL training.

Similarly, [71] employed an automated approach for designing fault diagnosis models using a combination of RL and NAS. Notable optimizations include a Greedy Strategy to prevent local optima, ER to smooth the learning process, and Weight Sharing to reduce computational demands. Moreover, the researchers employed a DQN-based RL framework to optimize a bandpass filter's upper and lower cutoff frequencies for fault diagnosis in rotating machinery signals in [147]. The bandpass filter acts as an agent, and its position, defined by these frequencies, represents its state. The agent interacts with the signal environment to maximize a reward signal based on how effectively it highlights fault characteristic frequencies. The DQN algorithm iteratively refines the frequency band to reveal the optimal range for fault identification. Experimental results on gear and bearing fault signals demonstrate that the proposed deep reinforcement learning-based method outperforms other traditional techniques, such as fast Kurtogram and GiniIndexgram, in identifying fault-related frequencies.Study [116] focused on developing an online domain adaptation learning method for fault diagnosis in machinery, which is achieved through the use of a Capsule Network (Cap-net) as an agent for autonomously extracting fault features from online data. A feature dictionary based on Coarse-grained Similarity (CS) is designed to label the online data, and a reward mechanism based on CS is used to evaluate the coarse-grained labels. The method consists of initializing the Cap-net, updating it with online data responses and rewards, and fine-tuning it with historical data. The target and evaluation networks are updated iteratively, and a self-pruning mechanism is employed to optimize the online feature dictionary.

Some other works employing RL in MFD are mentioned in Table 11.

**Table 11**
Articles employing RL employed in MFD.[7]

| AR | M | Dataset | Algorithm(s) | Result | Remarks |
|---|---|---|---|---|---|
| [70] | B, G | CWRU, UoC | FD as a guessing game; Classification using 1-D CNN-RL and GRU-RL | CNN-RL: CWRU=99.95%, UoC=99.61%; GRU-RL: CWRU=99.98%, UoC=99.95% | Actions are chosen via $\epsilon$-greedy method; ER updates Policy-Net |
| [129] | G | SEU | FD as a guessing game; DQN agent uses two 2-D CNNs with the same structure: Eval-Net (Updating) current Q value, Target-Net (Calculating fixed target Q value) | Overall accuracy > 99.5% | raw signals transformed into 2-D TF maps (agent's observation) using synchro-extracting transform; $\epsilon$-greedy exploration |
| [40] | B, G | CWRU, THU | FD as a guessing game; DDQN for solving class imbalance problem | Increment in class imbalance decreased the F1 scores | Higher rewards for diagnosing minority classes |
| [71] | B, G, BS | CWRU, PHM-2009, IMS, HOUDE, BS | Automatic NAS using RL; Creates CNN models for fault diagnosis tasks; Greedy Strategy to avoid local optima | 100% accuracy in test data | ER for smoother learning; Weight Sharing to reduce computational demands |
| [147] | B, G | MFS-SQ | DQN-based RL optimizes bandpass filter cutoff frequencies for FD; Bandpass filter acts as an agent, with frequencies defining its state. | Outperforms fast kurtogram, GiniIndexgram, and Smoothnessgram methods | DQN iteratively refines the frequency band for optimal fault identification |
| [116] | B | PU, MFPT | Utilized a Capsule Network for online domain adaptation; A reward mechanism based on CS evaluates labels; Self-pruning mechanism optimizes the online feature dictionary | Accuracy for different testing conditions are MFPT: Average accuracy 92% and PU: $\geq$ 96.25% | T-SNE for feature visualization in 2-D space;target and evaluation networks are updated iteratively |
| [114] | B | PU, JNU | Contrastive augmented DRL for fault classification in class imbalanced data; Two-stage training: Pretraining (CL to distinguish fault classes), fine-tuning (dynamic reward function for improved recognition of minority class) | 6-7% accuracy improvement than other models | Adaptive rewards improved stability & performance; Contrastive loss minimized overlap |

[7]BS:Ball Screw; G:Gears; LBD:Locomotive Bearing Dataset; TF:time-frequency

## 4.4. Transfer Learning-based FD

Transfer learning, a promising paradigm in ML, addresses the limitations of DL in practical applications, particularly when dealing with limited data and distribution discrepancies between training and testing datasets [21]. DL models typically require large amounts of labeled data to perform well, which is often impractical in industrial settings where collecting such data, especially faulty data, is time-consuming and challenging. TL mitigates this issue by transferring knowledge from related but different domains to improve model performance in new tasks, thus addressing data scarcity and reducing computational demands [372]. By utlizing knowledge from related domains, TL not only enhances learning with limited data but also ensures robust generalization and effective feature extraction from complex datasets. This makes TL a valuable tool in applications where traditional ML methods may struggle. In MFD, TL is widely applied, and for this review, we have categorized its use in this field into two categories:

### *4.4.1. Direct Transfer*

The direct transfer approach involves using pre-trained models developed on one dataset (source domain) and applying them to another dataset (target domain) with minimal or no additional training. This method is particularly useful when data from the target domain is scarce, as it allows the model to use patterns and features learned from similar but different conditions. The success of this approach depends on the similarity between the source and target domains; more similar domains lead to better performance. Although fine-tuning may be needed if the domains differ significantly, the direct transfer approach can effectively address the issue of labeled data scarcity. For instance, a model pre-trained on a large, labeled dataset like the CWRU-bearing dataset can be applied to a target domain with sparse data or different operating conditions. The model's performance is evaluated using metrics like accuracy and precision, and fine-tuning is conducted if necessary to ensure it adapts well to the new domain. Studies [45–47, 113, 373–375] are some of the works employing pre-trained models for MFD.

### *4.4.2. Domain Adaptation*

Recently, domain adaptation methods have been utilized widely in MFD, particularly when dealing with scenarios when there is a discrepancy between the source and target domains, Unlike direct transfer methods, which may suffer from significant performance degradation due to domain shift, DA methods are designed to mitigate these issues by focusing on reducing the distribution differences between domains. Domain adaptation relaxes the conventional assumption that training and testing data must be independent and identically distributed *(iid)*. By utilizing invariant features and essential structures across different but related domains, DA methods effectively address challenges such as domain shift, small sample sizes in the target domain, and unbalanced datasets [21, 372]. In this approach, a previously trained model is fitted to a new but related domain to minimize discrepancies between domains. The use of DA methods in deep TL can be subdivided into three categories: the discrepancy-based method, the adversarial-based method, and the reconstruction-based method, which are briefly described below.

1. **[A.] Discrepancy-based Methods:** The discrepancy-based domain adaptation method aims to minimize the differences between the source and target domains by reducing the discrepancy in the feature layer within a neural network. This method primarily focuses on aligning the feature distributions of both domains, enabling the model to generalize effectively to the target domain. In discrepancy-based DA, the core idea is to measure and reduce the distance between the source and target domains by applying statistical approaches to the feature layers of the model [372, 376]. Statistic transformation, structure optimization, and geometric transformation are key strategies for minimizing discrepancies between source and target domains in domain adaptation. By adjusting the statistical properties of feature distributions, such as mean and variance, techniques like Maximum Mean Discrepancy (MMD) and Correlation Alignment (CORAL) reduce domain differences. Structure optimization involves modifying the model's architecture or adding layers to enhance domain alignment. Additionally, geometric transformation aligns the feature space's geometric properties, ensuring consistency between the source and target domains, further reducing discrepancies. Studies [48, 49, 54, 56] implemented discrepancy-based DA approach in their work.
2. **[B.] Adversarial-based Methods:** Adversarial-based DA methods employ a domain discriminator to promote domain confusion and learn invariant features between source and target domains. Inspired by a two-player game as in GAN, this approach consists of a generator and a discriminator competing against each other, where a generator tries to create data that closely resembles the target domain, while the discriminator attempts to distinguish between the source and generated data. The adversarial training process involves the generator trying to confuse the discriminator, leading to better domain alignment. These methods are divided into Generative Adversarial DA (GADA), a method with an additional generator, and non-generative adversarial DA (non-GADA), a method without an additional generator. GADA utilizes a generator to produce data samples that resemble the target domain, aiding in domain alignment through transformations like those seen in Coupled GANs and adversarial discriminative domain adaptation (ADDA). Non-GADA, however, focuses on domain alignment without generating new samples, using the generator as a feature extractor to maximize domain confusion. Techniques like Wasserstein distance-based optimization, curriculum learning, and domain-symmetric networks are employed to reduce domain shift. While adversarial-based DA methods offer advantages like high sample diversity and theoretical proximity to target data, they face challenges with training stability due to difficulties in achieving Nash equilibrium [24, 372]. Studies [50, 51, 155, 377–380] employed adversarial-based DA method in their work.

3. **[C.] Reconstruction-based Methods:** Reconstuction-based DA method focuses on reducing domain differences by reconstructing the data from both source and target domains, ensuring intra-domain distinctions while capturing inter-domain commonalities. These methods typically use a coding-decoding framework, where data is encoded into feature representations and then decoded back to the original input, creating a shared domain space for alignment. By sharing an encoder, the model learns domain-invariant features while preserving domain-specific characteristics. Techniques like self-encoders and KL divergence enhance domain alignment, as seen in domain separation networks, which separate domain-invariant and domain-specific features to prevent negative transfer. Hybrid models combining AEs with GANs further improve performance by incorporating tasks like cyclic consistency loss, which is particularly useful in unsupervised scenarios. Despite the advantages, challenges remain in balancing the authenticity and diversity of generated samples and managing the complexity of parameter updates, making reconstruction-based DA a powerful yet intricate approach to domain alignment [21, 372]. Studies [52, 123, 150, 381] applied reconstruction-based techniques in their work.

#### *4.4.3. Multi-source Domain Adaptation*

Multi-source domain adaptation (MDA) uses data from multiple source domains with different distributions to improve model performance on a target domain. Unlike traditional single-domain adaptation, MDA addresses the challenge of conflicting information by aligning features across multiple domains, enhancing the model's generalization. Key approaches include residual adaptive modules for compressing and sharing parameters, latent domain discovery for finding commonalities, and domain weighting to select the most relevant source data. Adversarial training with GANs further strengthens MDA by minimizing distribution differences between domains. Despite its effectiveness, MDA still faces challenges in automatically selecting and aligning the most relevant data, making it a focal point for ongoing research in fault diagnosis [2, 372].

#### *4.4.4. Partial Transfer Learning*

Partial transfer learning is a specialized approach where only a subset of features or tasks from the source domain is relevant to the target domain [21]. In the context of MFD, this involves selectively transferring knowledge related to specific faults or operational conditions, rather than applying the entire model from the source domain. This method is particularly useful when the target domain has a narrower scope or when the source domain contains redundant or irrelevant information. By focusing only on the most relevant data, partial transfer learning minimizes the risk of negative transfer–where irrelevant knowledge could degrade model performance—-and ensures that the transferred information enhances the target model's accuracy and effectiveness. However, identifying the relevant subset requires careful analysis and domain expertise, making it a challenging but crucial aspect of successful fault diagnosis [381].

With the concepts of TL outlined, these approaches have been increasingly applied to address practical challenges in MFD. Guo et al. [98] addressed the challenge of limited labeled data and differing training/testing distributions by employing a deep convolutional TL network. This network comprises two modules: condition recognition and domain adaptation. The condition recognition module uses a 1-D CNN to automatically extract features from raw vibration signals and classify machine health conditions. The domain adaptation module further enhances the 1-D CNN by incorporating a domain classifier and a distribution discrepancy metric, enabling the model to learn domain-invariant features. This setup maximizes domain classification errors while minimizing the probability distribution distance between source and target domains. The approach was validated through six transfer fault diagnosis experiments on three different bearing datasets, showing significant improvements in fault diagnosis accuracy compared to traditional methods. Similarly, the study by Wei et al. [134] presented a method for machinery fault diagnosis of THU planetary gearbox dataset and UA-FS gearbox dataset. The methodology involved using the raw vibration signals and weighted domain adaptation network to address the challenge of data distribution shifts due to changing working conditions by assigning weights to conditions based on their similarity target. The study demonstrates improved diagnosis accuracy, utilizing classification accuracy and MMD as metrics, emphasizing the importance of domain adaptation in variable working conditions. Another study [79], proposed the wavelet packet transform (WPT)-based deep feature transfer learning method for bearing fault diagnosis under different working conditions, in which the combination of WPT (for time-frequency feature map construction), a deep ResNet (for features extraction), and a multi-kernel MMD (for evaluation of distribution differences in features across domains) is used. The study carried out on the CWRU bearing dataset, MFS-RDS (Rotor Dynamics Simulator) dataset provided average of 08.59% accuracy result in the CWRU dataset, whereas 97.14% accuracy in the MFS-RDS dataset. Talking about multi-source domain adaptation, the study [67] by Rezaeianjouybari et al. highlighted the limitations of previous models that rely on a single source domain and ignore variations in working conditions, and proposed feature-level and task-specific distribution alignment multi-source domain adaptation (FTD-MSDA) model to address the challenge of domain shift in intelligent fault diagnosis systems. Experimented on the CWRU and PU bearing dataset, the FTD-MSDA model framework aligned domains at both the feature and task levels, using Sliced Wasserstein discrepancy for shaping task-specific decision boundaries and successfully transferred knowledge from multiple labeled source domains to a single unlabeled target domain. Moreover, a study by Qian et al. [382], introduced the TL method called adaptive intermediate class-wise distribution alignment, targetting the fault diagnosis in the wind turbine planetary gearboxes, addressing the challenges like slow convergence and loss oscillation in DA. The proposed model, employing both domain adaptation and generalization, utilized an adaptive intermediate distribution mechanism alongside an AdaSoftmax loss, which dynamically aligns the source and target domain distributions without requiring additional losses for distribution distance or correlation regularization. Furthermore, a study by [383] developed the variance discrepancy representation method to enhance the DA in FD of rotaing machinery. The experimental results across three bearing datasets demonstrated that the proposed method improved the representation of distribution discrepancies by focusing on variance rather than mean, providing a more accurate reflection of differences between source and target domains.

**Table 12**
Articles employing TL in MFD. [8]

| AR | M | Dataset | Algorithm(s) | Result | Remarks |
|---|---|---|---|---|---|
| [89] | B | CWRU, PU | Transferable decoupling multi-scale AE: Multi-scale residual network with transposed convolution for distribution alignment, feature extraction, classification, and domain adaptation | Achieved upto 100% in domain transfer tasks in certain sub-datasets | PP: Data cleaning and segmentation |
| [384] | B | QPZZ-II | Joint distribution adaptive DBN optimized with a Sparrow Search Algorithm variant | Average accuracy 79.21%, peak accuracy 90.71% in specific conditions | PP: Time-domain data converted to frequency-domain via FFT |
| [385] | G | WT-DDS | Deep TL integrating cohesion-based sensitive feature selection, a three-layer SpAE for feature extraction, MMD for aligning data distributions across different datasets, and SoftMax as classifier | Highest classification accuracy of 99.17% | Signals from four sensors averaged over three experiments |
| [386] | B, G | CWRU, Gear Fault Dataset | Integration of unsupervised geodesic flow kernel (GFK)-based DA with Z-normalization and SoftMax regression | Average detection of 99.9% | PP: STFT to raw vibration data, converting it to a power spectral density matrix in decibel form, and then flattening it into a 1-D vector for GFK algorithm compatibility |

[8]WF: Wind farm; WT-DDS: Wind Turbine Drivetrain Diagnostics Simulator; WTG: Wind turbine gearbox

| | | | | | |
|---|---|---|---|---|---|
| [97] | B | CWRU, Self-built | Deep imbalanced domain adaptation, integrating empirical risk minimization, cost-sensitive re-weighting (class-balanced loss), categorical alignment (local MMD), and margin loss regularization (label-distribution-aware margin loss) | Average accuracy of 96.55% | Multi-channel vibration signal as input to 1-D ResNet, used as backbone |
| [387] | B, WTG | CWRU, WF Dataset | Instance-based deep boosted TL, utilizing algorithms like AdaBoost and TrAdaBoost for weight adjustments to minimize negative transfer effects | 9.79% improvement from base case | PP: Noise removal; DataAug:Sliding window technique |
| [388] | B | CWRU, self-built | Self-attention lightweight CNN and TL | Significant reduction in model parameter | PP: CWT |

## 4.5. Other Methods

### 4.5.1. MFD as Anomaly Detection

In the field of MFD, where the lack of labeled fault data is a challenge, un/semi-supervised AD techniques play an important role in identifying faulty patterns and potential anomalies that cannot be captured by traditional supervised methods. These data-driven approaches focus on detecting outliers or anomalies (data points that deviate significantly from the majority of normal data instances) [389]. Various classical unsupervised methods, such as Z-Score (measures how many standard deviations a data point is away from the mean), Interquartile Range (identifies outliers by considering the range between the first and third quartiles of the data), isolation forest (iF) (a tree-based algorithm that isolates anomalies by constructing isolation trees by measuring the number of splits required to isolate a data point), Local Outlier Factor (LOF) (measures the local density deviation of a data point with respect to its neighbors), One-Class SVM (learns a boundary around normal data instances and classify anything outside this boundary as an anomaly), and so on, are used to identify anomalies in datasets. Apart from these shallow learning, DL-based approaches have been used for AD in different ways, like feature extractors, representations learners from normal data, and end-to-end anomaly score learners. More details can be found in [389]. Advanced architectures like AE, VAE, and GAN are used mostly as AD algorithms. A study by [390] introduced the full graph dynamic AE (FGDAE)-based AD method, designed to operate effectively under complex and varied conditions. The FGDAE model integrated a fully connected graph to capture global structural relationships between sensor channels, a graph adaptive AE that aggregates multi-perspective features and adapts to changes in operating conditions, and a dynamic weight optimization strategy to handle training with unbalanced multi-condition data. Similarly, a study [128] proposed fault-attention generative probabilistic adversarial AE (FGPAA) method for AD of three machinery fault datasets by focusing only on healthy classes. Utilizing dual adversarial AEs, the FGPAA method employed a fault-attention probability distribution to evaluate the health state of machinery effectively, allowing for dynamic adjustment to signal noise and anomaly detection in real time. Moreover, a work by [391] proposed a real-time AD method for the fault diagnosis of marine machinery. The authors develop a framework named RADIS, based on LSTM-based VAE combined with multi-level Otsu's image thresholding technique. Furthermore, a GAN-based AD method was proposed in [357], where an encoder-decoder-encoder architecture was employed in the generator to train the normal samples exclusively. The latent and apparent losses to compute anomaly scores.

### 4.5.2. Use of Transformers

The Transformer, initially proposed by Vaswani et al. [392], revolutionized natural language processing and has been successfully applied in other areas like computer vision. These models, characterized by layers of Transformer blocks featuring multi-head self-attention and batch normalization, efficiently handle tasks without complex operations like CNNs or RNNs, often outperforming them. Recently, the use of architectures has been seen in MFD as well. A study [78] introduced a time-series transformer (TST) model for direct 1-D raw vibration data processing, without any signal preprocessing. TST leverages multi-head self-attention and transformer blocks for feature extraction from bearings and gear faults. Evaluation on the CWRU, XJTU, and UoC datasets shows impressive accuracy, e.g., 98.63% for CWRU and 99.78% for XJTU and 99.51% for UoC. TST's feature vectors exhibit superior intra-class compactness and inter-class separability, emphasized by t-SNE visualization. Similarly, a window-based multi-head self-attention model is proposed in [77], employing three datasets: CWRU, UoC, and Shandong University (SDU) dataset. Data preprocessing includes 1024 sample lengths and dataset partitioning. The model combines self-attention and CNN with a 1-D window-based multi-head self-attention for local feature learning. Results demonstrate the model's superior classification performance, achieving nearly 99.99% accuracy in noise-free conditions and maintaining robustness with added noise (SNRs from -6 dB to 6 dB). Moreover, an article by Wu et al. introduced a transformer-based classifier for machinery fault classification [393]. Employing CWT for the generation of time-frequency spectrogram images from raw data as an input, the authors employed a Mahalanobis distance-based technique to identify previously unseen faults.

#### 4.5.3. Physics Informed Neural Networks (PINN)

PINNs are an innovative approach in ML, which incorporates physical laws into the structure of neural networks. This integration enhances predictive accuracy and interpretability. In the domain of MFD, PINNs offer a noteworthy advancement by enabling precise identification and analysis of faults in mechanical systems. A study [394] employed PINN, applying physics-informed loss functions, to enhance the model's interpretability for fault severity identification of axial piston pumps. The research utilized a high sampling rate to collect data on axial piston pumps, identifying piston wear in four severity levels using a low-pass filter to isolate relevant frequencies. The proposed PINN model accurately identified wear states by estimating gap clearances linked to the pump's health indicators. Similarly, researchers studied early fatigue in the main bearings of wind turbines and aimed to predict the RUL of these bearings using PINN method in [395]. This model assessed bearing fatigue and grease degradation using data such as wind speed, bearing temperature, and grease analysis from a 1.5 MW wind turbine. Moreover, Ni et al. [396] implemented a physics-informed residual network for fault diagnosis using a rolling element bearing dataset from a drivetrain diagnostics simulator. The network featured a modal-property-dominant layer with cepstrum exponential filtering to emphasize system properties, a domain-conversion layer using Computed Order Tracking to address speed variations, and a parallel bi-channel architecture for extracting complex fault features, resulting in improved diagnostic accuracy under varying conditions.

#### 4.5.4. Federated Learning

Federated Learning (FL) introduces a decentralized model training method, especially useful in edge learning environments where data is distributed across multiple devices. It trains a shared model via iterative updates using local data, avoiding the centralization of sensitive information. This technique facilitates collaborative training while safeguarding data privacy and minimizing communication overhead. FL has been widely adopted for edge ML models and successfully applied in areas like cyber attack detection, spam detection, smart cities, autonomous vehicles, and recently machinery fault diagnosis, primarily aiming to maintain privacy [397]. In a study by [398], the researchers implemented an FL-based method for detecting mixed faults, particularly focusing on the rotor and bearing. The input method is 1-D time-series data collected from accelerometers, with a dataset comprising 92,160 points across 48 mixed fault classes. The methodology includes an FL approach utilizing a duplet classifier, where data is distributed among 30 clients in three partition schemes: balanced *iid*, balanced non-*iid*, and unbalanced non-*iid*. Preprocessing involves data shuffling and partitioning. The FL model, trained with CNNs, achieved over 90% classification accuracy for mixed faults. Another study by [399] focuses on improving fault diagnosis in rolling bearings using federated learning. Using data from the CWRU benchmark dataset and a project dataset from Nio Inc., the study aims to address client inhomogeneity issues (related to sample size, quality, and fault type) in FL. The proposed methodology is a multi-scale layer-by-layer recursive fusion federated learning method (LLRFed). Data is preprocessed using a sliding window and fed into a DNN with FFT. The study demonstrates that the LLRFed method significantly improves diagnosis accuracy by 9.23% accuracy in the benchmark dataset. Similarly, the fault diagnosis method for railway point machines used in high-speed and rail systems is proposed in [400], where the vibration signals from these machines under various fault conditions were analyzed using a sequential and asynchronous FL framework. The dataset included 960 vibration signal samples across sixteen conditions, each with 60 samples, and was preprocessed to ensure consistent 10-second vector lengths using trimming and zero-padding. The study employed a deep shrinkage fully convolutional network as the global model, reducing parameters by one-fourth compared to previous models, thereby minimizing communication expenses and packet loss. Data were split into training and testing sets (8:2 ratio), with one-hot encoding applied to labels. Similarly, a study by Yang et al. [401] employed an FL approach to transfer the diagnosis of industrial machines, such as bearings and robots. Their TL through distribution barycenter medium architecture integrates an FL learning framework with a server-client architecture to address data decentralization challenges, where data from different domains cannot be centrally aggregated due to privacy concerns and high transmission costs. Clients construct ResNet-based diagnosis models to represent high-level features from local data, while the server implements an StAE-based generator model to produce a distribution barycenter medium by aggregating domain-specific distribution parameters. Using balanced samples of vibration data across different health states, this method adapts marginal and conditional distributions through the generated medium samples and employs collaborative training between clients and the server to dynamically update the distribution barycenter medium. Furthermore, the study [402] presented a blockchain-based decentralized collaborative learning approach for machine fault diagnosis.1-D machinery data from two datasets, a high-speed train bogie and a shaft crack failure dataset, were used. Deep CNNs are used for the analysis in a framework that combines blockchain-based federated learning with source data-independent transition learning, in which pre-processing techniques include the use of frequency-domain information. This approach achieved over 90% testing accuracy. Moreover, the article [403] presented the use of a heterogeneous federated domain generalization network, which incorporates common representation learning to address the challenges of domain shift and privacy preservation in FD. Implementing the federated TL to achieve generalized FD across different and unseen target clients by using heterogeneous source clients, the proposed method he;ped overcome the limitations of traditional methods that rely on homogeneity among clients and the availability of target-domain data during training. The model uses a disentangled domain adaptation base model designed to minimize noise impact and enhance domain confusion, improving the extraction of fault-relevant features.

## 5. Challenges

In the field of MFD, despite significant academic attention, challenges remain for effectively applying advanced data-driven algorithms to real-world applications. These challenges encompass various aspects, such as dataset issues, model architecture, and existing approaches, which are elaborated upon in the subsequent subsections.

### 5.1. Data related challenges

For employing the advanced learning approaches for MFD, a huge amount of data is needed. The available datasets, however, have many limitations which restrict the employed models from performing well. Listed are the challenges the researchers face when working with machinery fault datasets:

1. **Challenges related to sensors**
    - **Smart sensors:** The various challenges associated with sensor networks include sensor fusion, security and privacy, network traffic, and energy efficiency. Sensor fusion presents the difficulty of accurately integrating data from multiple sensors into a single output, necessitating the careful selection of peripherals to achieve high efficiency while minimizing power consumption and noise. Ensuring security and privacy is a crucial challenge, particularly in protecting data within organizational boundaries and in cloud computing contexts. Network traffic is another significant challenge, as the simultaneous data transmission from multiple sensors can result in increased network congestion and the risk of data loss. Lastly, energy efficiency poses a challenge in minimizing power consumption while optimizing resource utilization across the sensor network.
    - **Multi-sensor data fusion:** Challenges in multi-sensor data fusion include managing data complexity, handling imprecision and uncertainty, integrating both homogeneous and heterogeneous sensor data, efficiently analyzing unrelated data from diverse sources, synchronizing data in distributed systems to prevent real-time performance issues, and balancing the high costs of monitoring frameworks and data acquisition systems with the computational decisions between using edge or cloud computing.
2. **Problems with the Real-world data:** Real-world machinery data is often noisy, inconsistent, come from heterogenous sensors, is non-stationary, and incomplete, making fault analysis challenging. This can result from sensor failures, intermittent faults, or communication issues during data collection [236]. The contrast in data quality between real-world scenarios and controlled lab environments presents significant challenges for practical applications. Furthermore, the reluctance of industries to provide real industrial datasets for research is also an existing problem.

3. **Faulty Data Scarcity:** Real-world faulty machinery data is often scarce, making it challenging to train advanced ML models. Obtaining an adequate amount of faulty data for machinery fault detection is challenging because machinery typically operates under normal conditions. Again, the faults are of different types and severity, and obtaining enough data for each type of fault is a challenging task [1].
4. **Insufficient Labeled Data:** As previously mentioned, the majority of works rely on supervised learning, necessitating labeled data. However, precisely labeling fault types and severities is challenging, contributing to the scarcity of labeled datasets. This shortage of labeled data is a significant challenge in real-world industrial contexts, as labeling is a costly and time-consuming process, often lacking diversity to ensure effective generalization to unseen faults [25].
5. **Data Imbalance:** The faulty data scarcity leads to a class imbalance problem, with more data on normal states than failures, leading to the under-representation of certain fault types. This imbalance can hinder the learning process and bias the model towards the majority class [404].
6. **Data Incompatibility:** Detecting machinery faults becomes challenging when data comes from different sources, each with its own objectives, complexity, and criteria for data handling. Additionally, variations in data storage depth within databases create modeling issues for fault detection in machinery [405].
7. **Transferring Knowledge from Research Labs to Practical Applications:** Most of the research works are carried out in the publicly available datasets, which are acquired from the lab in a controlled environment, aiming to apply this knowledge to detect unseen machinery faults, including predicting real-world faults from lab-generated data. However, several technical challenges must be addressed to achieve this goal [8].
8. **Problems with the available Datasets:** Most of the machinery fault datasets have limitations, some of which are listed below:
   - Diverse challenges arise when using publically available datasets in MFD. These challenges include non-classical fault recognition features, non-stationary characteristics, difficulty in identifying all faults accurately, high variance, data corruption, irregular frequency components, missing values, limited representation of fault types, restricted accessibility, and a limited range of operating conditions [8, 13, 21, 43].
   - Additionally, some datasets contain heterogeneous data from various sources, which can hinder consistency and model training. Also, some are notably complex and challenging to analyze [33].
   - Many of these datasets are relatively small, which can result in limitations on model performance. Additionally, some datasets may contain missing data and potential biases toward certain labels or classes, further limiting model performance [1].

### 5.2. Challenges encountered in Rotating Machinery

- **Complex Movement:** Real-world machinery often involves sliding between components, which complicates the calculation of fault frequencies and affects feature informativeness.
- **Frequency Interference:** In cases where multiple types of machinery faults occur simultaneously, their interaction can lead to complex frequency interplay, making informative frequencies less clear.
- **External Noise:** Additional sources of vibration or acoustic emission or other types of noise can introduce interference and obscure relevant features.
- **Challenging Fault Types:** Some machinery faults can exhibit non-stationary behavior, lacking characteristic cyclic frequencies, making them challenging to detect.
- **Sensitivity Variability:** Features related to machinery defects can be sensitive to different operating conditions, requiring systematic adaptation.
- **Sensor Placement** Obtaining accurate machinry fault signals necessitates expensive sensors and expert involvement in securely mounting them onto machines. Inadequate data quality resulting from improper sensor placement can adversely affect the performance of DL models in fault detection [406].

### 5.3. Challenges in Existing Approaches

1. **Supervised Learning:** The predominant approach in MFD is the supervised learning technique. However, there are certain challenges, which are mentioned as follows:
   - These methods excel in detecting known faults but struggle with new or unseen types.
   - Acquiring a diverse, well-labeled, and balanced dataset for supervised methods is resource-intensive and time-consuming, frequently resulting in limited model generalization.
   - Feature selection is crucial in supervised learning, impacting model performance. It can be challenging with high-dimensional data.
   - Supervised learning can lead to overfitting, especially when there is limited training data, causing poor performance on new data.
   - Supervised learning algorithms' sensitivity to data noise can lead to false positives and false negatives, impacting fault detection accuracy.
2. **Semi-Supervised Learning:** The use of semi-supervised learning in MFD is increasing. They overcome the limitations of supervised learning techniques up to some extent; however, there are certain limitations, which need to be addressed.
   - Semi-supervised learning needs less labeled data than supervised methods but still faces challenges in obtaining quality labels, which can be costly in MFD.
   - Challenging to determine the optimal amount of labeled data, as too little leads to poor performance, and too much can cause overfitting.
   - Combining labeled and unlabeled data consistently can introduce biases and is not always straightforward.
   - Like supervised learning, managing class imbalance can be a challenge in semi-supervised learning, where labeled samples in each class may be imbalanced, potentially biasing models.
   - Complex implementation due to the need for expertise in both supervised and unsupervised domains.
   - Sensitivity to the quality of labeled data can degrade algorithm performance.
   - Difficulty in selecting the appropriate semi-supervised learning algorithm from a variety of options.
3. **Unsupervised Learning:** The challenges in employing unsupervised learning approaches in MFD are as follows:
   - The biggest advantage of unsupervised technique, i.e., not requiring labelled data, can be its limitation, as it can be difficult to evaluate the performance of the model.
   - Unsupervised techniques rely on detecting anomalies, which can be difficult with complex and diverse operational behavior.
   - Setting appropriate thresholds for anomaly detection is subjective and may require expertise.
   - These methods might generate false positives when processing complex machinery data, affecting reliability.
   - Unsupervised methods may lack clear result explanations, hindering precise fault diagnosis.
4. **Reinforcement Learning:** The limitations of RL in MFD are listed as follows:
   - Limited application of RL in MFD despite its success in other domains.
   - Underutilization of RL's potential in optimizing maintenance decisions and fault detection in existing systems.
   - Current RL algorithms for MFD often oversimplify fault diagnosis as a guessing game, resembling a classification task [40].
   - Limitations of RL algorithms in adapting to various machinery or environmental settings [116].
   - Challenges in balancing exploration and exploitation to discover optimal policies and maximize rewards in RL [364].
   - RL often requires extensive interactions with the environment, making generalization to unseen environments or tasks difficult [116].
   - Difficulties faced by RL in handling continuous action and high-dimensional state spaces.

### 5.4. Challenges in employing ML/DL Algorithms

ML/DL algorithms have made significant strides in mapping one space, $X$, to another, $Y$, using continuous geometric transformations, particularly when a large amount of data is available. This achievement has been revolutionary across various industries [407]. Yet, human-level artificial intelligence remains an elusive goal. Challenges researchers face with deeper networks include:

1. **Feature Engineering Challenge for Classical ML:** In MFD, classical ML algorithms rely on feature engineering to identify relevant fault indicators based on operational parameters and physical characteristics. However, this approach might face challenges that affect fault classification accuracy [8].
2. **Complex Model Training:** Training ML/DL models for machinery fault detection demands substantial computational resources and time, with deep neural networks being especially data-intensive.
3. **Model Interpretability:** Deep learning models are often regarded as **black box** models, and lack transparency and interpretability, which is a concern for safety-critical applications like machinery fault detection. Ensuring trust and facilitating decision-making in fault detection systems is crucial, necessitating research to enhance the interpretability of DL models in this context [408, 409].
4. **Generalization:** Most techniques focus on specific situations, not an integrated engineering environment, affecting their generality. Also, the existing algorithms often face challenges in finding the optimal balance between approximation and generalization. Attempting to approximate the training data too closely can result in overfitting while compromising the fit for the sake of generalization can lead to significantly poor performance on the training data [408, 410].
5. **Data Dependent:** The DL models are extremely data dependent and need enormous data to learn adequately. The DL models cannot perform well in situations in which there are not enough data. Again, DL models do not work well when the dataset is unbalanced. It misclassifies the class with a higher number of data samples. Moreover, DL models face challenges when learning from poor quality and redundant data in the context of MFD [411, 412].
6. **Computational and Expensive:** These models are computationally intensive and require powerful hardware for training and executing, which makes them expensive. This can be a limiting factor for small research teams and industrial applications with limited resources[1].
7. **Feature Extraction Challenges:** Selecting relevant and important features from raw sensor data is vital for accurate fault detection, but identifying the most informative ones within a large dataset can be complex. Effective feature extraction methods and domain expertise are required to extract meaningful features that capture fault-related patterns.
8. **No right way in selecting DL Architectures:** Selecting the appropriate DL tool and architecture for machinery fault detection is challenging due to the absence of standardized guidelines. DL models come with numerous hyper-parameters, demanding expert knowledge and computational resources for optimal performance. Additionally, most existing DL architectures are designed for image data, so engineers are encouraged to explore custom convolutional architectures tailored to industrial mechanical data's unique characteristics, holding potential for improved fault diagnosis [413]. The limitations of some of the mostly used DL architectures in MFd are listed below:
   - **Autoencoders:** The traditional autoencoders often include a pre-training stage to initialize the network weights, and errors in the first layers can potentially lead to the network learning to reconstruct the average of the training data. Also, AE has limited interpretability as the models typically learn a compressed representation.
   - **CNN:** CNNs rely on labelled data and require several layers to find the entire hierarchy. Also, CNNs face challenges in capturing long-term dependencies in time series data, have fixed input size constraints that pose problems for varying-length time series, are highly sensitive to hyperparameter tuning, and have limited ability to handle irregularly sampled data or missing values.
   - **DBN:** DBNs have limitations, such as unclear optimization steps based on maximum likelihood training approximation and their inability to account for the two-dimensional structure of input images, impacting their performance and applicability in computer vision and multimedia analysis tasks [414].
   - **GAN:** GANs may encounter instability and challenges in training convergence, are susceptible to mode collapse where the generator produces limited types of samples, face difficulties in evaluating the quality and performance of GAN-generated samples, and require careful tuning of the architecture and training process [415, 416].

- **RNN and LSTM:** These models face difficulties in training on long sequences due to vanishing/exploding gradients, exhibit a limited ability to capture complex temporal dependencies, and present challenges in parallelizing the training process.
- **TL:**
  - Transfer learning can help reduce data-related challenges but may not always be appropriate for MFD applications due to differences in operating conditions, sensor configurations, and machine types which may limit the effectiveness of knowledge transfer from one task to another.
  - Difficulty in improving diagnosis accuracy due to various factors such as transfer errors between different working conditions, unique characteristics of different components, scale amplification issues between laboratory and industrial fields, and limitations imposed by sample size and prior knowledge.
  - Negative transfer in transfer learning leads to poor performance in the target task. Preventing negative transfer is still an open question, and more negative transfers occur if the source task significantly differs from the target task. Selecting source data that is more similar to the target data can help reduce negative transfers [21].
  - Developing fault detection models that can be applied across different machines or equipment types is challenging. Each machine has unique operating characteristics, environmental conditions, and fault manifestations. Ensuring the generalizability of the fault detection models to new, unseen machines remains a significant challenge.
  - Though MFD has made significant advancements in the areas of fault recognition, extraction of fault features, dynamic condition monitoring, evaluation of fault severity, and prediction of RUL, current research indicates that deep transfer learning techniques have primarily been utilized for fault feature extraction and
  - Computational burden arising from additional complexity during transfer between source and target domains, as well as the inherent computational demands of deep learning architectures.

### 5.5. Other Challenges

- Lack of cross-validation and experiment repetitions in most of the research.
- Insufficient explanation of pseudo-code in proposed methods.
- Selecting appropriate evaluation metrics for fault detection models is crucial. Choosing metrics that reflect the real-world impact of false positives and false negatives can be complex.
- Properly performing cross-validation with machinery fault data can be challenging due to the temporal nature of the data and the need to ensure that no data leakage occurs.
- In some industries, there may be regulatory requirements for implementing and validating fault detection systems, adding an extra layer of complexity.
- Implementing fault detection systems across an entire industrial facility with numerous machines may pose scalability challenges in terms of hardware and infrastructure.

## 6. Recommendations for Future researchers

### 6.1. Machinery and Dataset Enhancements

1. **Study and Analysis:** As the initial and foremost step, we recommend that future researchers have a thorough and detailed study of the dataset they will be working on. As mentioned multiple times, data is fundamental for the performance of any algorithm. Thus, the careful analysis of the data prior to the development of the model might help in a better understanding of the problem and performance of the model.

2. **Dataset Creation:** When generating a dataset, careful considerations should be made.
    - Incorporate of high-quality sensors like accelerometers and acoustic emission sensors for data acquisition under varying conditions (healthy and faulty states, different severities).
    - Proper sensor placement is essential for accurate data capture, requiring sensors to be positioned close to the source and securely mounted to minimize interference and noise [160].
    - est-beds should be designed in such a way that they replicate real-world conditions, with carefully chosen materials and configurations.
    - Collect data across different load and speed settings, ensuring it captures meaningful fault signatures.
    - Label datasets accurately with details on fault types, severities, and experimental conditions.
    - Apply preprocessing techniques like filtering and normalization to refine data for algorithm development.
    - Organize data systematically for easy access and analysis. If sharing publicly, host datasets on reliable platforms with comprehensive documentation to support other researchers in the field.
3. **Sensor Strategy:** Adjust the number of sensors based on the chosen method. Classical methods may require fewer sensors, while deep learning approaches benefit from multiple sensors at various locations. Integrate data from diverse sources, such as vibration, temperature, and acoustic signals, to enhance fault pattern recognition and model performance [32].
4. **Dataset Expansion:** Ensure adequate dataset size for training robust deep learning models. Employ data augmentation, SP techniques, and synthetic data generation to increase dataset diversity. Techniques like transfer learning and domain adaptation can mitigate issues related to data scarcity [405].
5. **Data Preprocessing:** Implement data preprocessing to remove noise, address missing data, and filter out irrelevant information. Convert raw data into formats like frequency or time-frequency domains to expose hidden patterns, thus boosting the efficacy of fault detection algorithms.
6. **Comprehensive Model Evaluation:** Use a mix of common benchmarks and real-world industrial data for a thorough evaluation of models, ensuring they are tested under diverse conditions and can handle practical challenges in MFD.
7. **Data Fusion and Handling:**
    - Enhance data collection through multi-sensor data fusion for comprehensive multi-fault diagnosis.
    - Apply cost-sensitive learning and ensemble methods with resampling strategies to manage data distribution.
    - Use datasets that reflect the imbalanced nature of real industrial settings for robust model training.

## 6.2. Algorithms Development:

- Assessing the working environment and operating conditions properly is another recommendation. For simpler setups, classical ML methods or frequency-based models are suitable. In noisy or complex environments with multiple operating points, advanced DL approaches are recommended, incorporating denoising techniques for noise resilience.
- Some studies [417, 418] suggest that deep learning isn't always superior to traditional methods in MHM. It's advisable to begin problem-solving with simpler methods.
- Select an appropriate DL/ML architecture that balances complexity, interpretability, and computational requirements for a specific MFD task [24].
- The use of unsupervised-based AD methods can solve the labeled data scarcity and class imbalance problem. The faulty data are anomalous or different in pattern, and thus, the anomaly detection methods detect them without the need for labels. Also, the new and unseen faults are also expected to be detected.
- Use of semi-supervised or unsupervised learning algorithms can be one solution for labeled data scarcity. When using SSL, employ an appropriate proportion of labeled data while emphasizing the integration of unlabeled instances.

- Apart from data augmentation techniques, the data imbalance problem can also be addressed through the design of loss functions. For example, the Pareto optimization-based approach in [419] introduced a multi-task learning framework that balanced the trade-offs between tasks to handled data imbalance effectively. Similarly, the adaptive cross entropy loss was designed to automatically reweight the influence of majority and minority samples during training. By down-weighting well-classified majority samples and focusing on the minority samples that are difficult to classify, this approach enhances the model's ability to handle imbalanced data effectively.
- Apply regularisation techniques and denoising methods to clean signals and avoid overfitting and improve the generalization performance of DL models in MFD.
- Conduct comprehensive testing using cross-validation methods and repeated experiments to validate model performance, stability, and generalizability.
- Integrate DL with traditional ML, SP, feature engineering, and other techniques to leverage their combined strengths for a robust MFD algorithm [166].
- Investigate domain adaptation methods to improve the effectiveness of transfer learning in MFD applications, considering different operating conditions and machine configurations.
- On the application of RL, explore the development of new reward functions, algorithms, and learning paradigms more suited to MFD applications to address the particular challenges of machine fault detection.
- Incorporate explainable AI techniques to enhance the interpretability of deep learning models, offering insights into their decision-making processes, and increasing confidence in their predictions.

### 6.3. Other Recommendations:

- Extend the research scope to include a wider range of faults and machinery. While existing studies predominantly focus on specific machine components like bearings, gears, and motors, real-world applications frequently involve these components operating together. This necessitates the development of diagnostic methods that can assess and understand the interactions between various machine parts within combined systems.
- To enhance the accessibility and comparability of methods, it is highly recommended to publish the source code of your machinery fault detection algorithms. Providing open-source code allows other researchers and practitioners to implement, evaluate, and build upon your work. This transparency contributes to the advancement of the field and encourages collaboration and innovation.
- The authors also suggest that researchers should adopt a thorough and structured documentation approach, which can benefit future researchers in the field.
- Consider incorporating online learning techniques for MFD to enable continuous adaptation to changing machine conditions and new failure types [116]. Employing digital twins is beneficial for simulating scenarios to understand root causes of failures, predicting faults for proactive maintenance, and enabling real-time monitoring [317].
- Although there is a lack of availability of open-source codes in this field, there are a few researchers who have maintained open-source resources. We recommend that future researchers visit their sites for assistance: `https://qinyi-team.github.io/#blog`, `https://github.com/liguge`, `https://biswajitsahoo1111.github.io/`, and so on. We also urge future researchers to make their code and datasets open-source, as this would greatly benefit the entire community working in this field.

## 7. Future Prospects

The future of MFD holds several promising opportunities, including developments aimed at enhancing the accuracy, dependability, and adaptability of diagnostic systems in industrial settings. First, researchers are focusing on improving the robustness of fault detection and classification by investigating new methods to detect crack damage under mixed-speed conditions. This research highlights the importance of noise-resilient algorithms, which are essential for accurate diagnosis in noisy industrial environments. Furthermore, the integration of advanced models such as transformer architectures and PINNs, which incorporate physical laws to enhance learning accuracy, is set to advance MFD systems. Researchers are also exploring various data fusion methods—including sensor, feature, and decision fusion—to merge different data types like vibration, current, torque, acoustic, visual images, and so on with machine learning. This approach optimizes the preprocessing of these data types, improving the accuracy and efficiency of fault detection. Thus, the development of methods to handle complexity between different fusion levels is the prospect future reserach.

The exploration of semi-supervised and reinforcement learning in MFD is emerging. focusing on reducing dependence on labeled data and employing sequence learning to predict faults and enhance detection efficiency. Future research in MFD is increasingly emphasizing the practical applications of offline RL aiming for better generalization in new environments. This shift prioritizes detecting faults over simply classifying them, reflecting a more dynamic and proactive approach to fault management.

Furthermore, the integration of advanced techniques like domain generalization, domain confusion, and domain adaptation is set to address unforeseen machinery fault conditions, data imbalance, and limited labeled data. Also, the development of XAI approaches promises to make MFD more transparent and interpretable, building trust in the decision-making processes of these systems. Overall, the future of machine fault detection research is characterized by innovation and practicality, promising more reliable, accurate, and adaptive fault detection methods for industrial applications.

Last, but not the least, to tackle the computational challenges of real-time diagnostics, refining data handling capabilities and computational efficiency is crucial, employing methods such as pre-training, NAS, and model compression. Integrating mechanism knowledge with DL is expected to improve both model interpretability and generalization, leading to more autonomous and sophisticated monitoring systems. The incorporation of digital twin technology and hybrid data-driven approaches will also play a significant role in minimizing prediction errors and refining maintenance strategies. Collectively, these are the future research possibilities, which will advance MFD not only to meet the complex demands of modern industries but also to advance toward more reliable and efficient fault detection methodologies.

## 8. Conclusion

This review offers a comprehensive overview of machinery fault diagnosis approaches, encompassing a wide range of topics from data sources to maintenance techniques, and predictive maintenance strategies. It provides an extensive exploration of available datasets in this field and covers traditional approaches as well as advanced methodologies, including deep learning, federated learning, reinforcement learning, transfer learning, and physics-informed neural networks. The review emphasizes the critical importance of data collection and maintenance analysis, highlighting the potential of various predictive methods to enhance fault detection and machine reliability. The challenges identified serve as key areas for future research, while the recommendations provided aim to improve dataset quality, algorithm selection, and practical application considerations. This review is a valuable resource for both newcomers and experts in the field of MFD, and we believe it will help to improve fault detection and machine reliability in a variety of industries.

# 9. Appendix

## Appendix A: Abbreviations used in the article and their full forms

**Table 13**
Abbreviations and their Full Forms

| Abbreviation | Full Form | Abbreviation | Full Form |
|---|---|---|---|
| AcE | Acoustic Emission | ACGANL | Auxiliary Classifier Generative Adversarial Network |
| AD | Anomaly Detection | AE | Autoencoder |
| ANN | Artificial Neural Network | AT | Automobile Transmission |
| AUC | Area Under the Curve | BEGAN | Boundary Equilibrium Generative Adversarial Network |
| BiRNN | Bidirectional Recurrent Neural Network | BiGRU | Bidirectional Gated Recurrent Units |
| BPNN | Backpropagation Neural Network | CAE | Contractive AE |
| Cap-Net | Capsule Network | CbM | Condition-based Maintenance |
| CDCN | Coupled Dense Convolutional Network | C-MAPSS | Commercial Modular Aero-Propulsion System Simulation |
| CNN | Convolutional Neural Network | CNNFD | CNN-based Fault Diagnosis |
| CGAN | Conditional Generative Adversarial Network | ConvAE | Convolutional AE |
| CPF | Chemical Process Faults | CRNN | Convolution Recurrent Neural Network |
| CS | Coarse-grained Similarity | CWRU | Case Reserve Western University |
| CWT | Continuous Wavelet Transform | DAE | Denoising AE |
| DBN | Deep Belief Network | DCNN | Deep CNN |
| DCGAN | Deep Convolutional Generative Adversarial Network | DFT | Discrete Fourier Transform |
| DIRG | Dynamic and Identification Research Group | DL | Deep Learning |
| DNN | Deep Neural Network | DQN | Deep Q-Networks |
| DRL | Deep Reinforcement Learning | DT | Decision Tree |
| DSTG | Defence Science and Technology Group | EBGAN | Energy-Based Generative Adversarial Network |
| EMD | Empirical Mode Decomposition | ER | Experience Replay |
| FD | Fault Diagnosis | FEMTO-ST | Franche-Comté Électronique Mécanique, Thermique et Optique - Sciences et Technologies |
| FFT | Fast Fourier Transforms | FGDAE | Full Graph Dynamic AE |
| FGPAA | Fault-Attention Generative Probabilistic Adversarial AE | FTD-MSDA | Feature-level and Task-specific Distribution Alignment Multisource Domain Adaptation |
| GAN | Generative Adversarial Network | GRU | Gated Recurrent Units |
| HHT | Hilbert-Huang Transform | HMM | Hidden Markov Model |
| HTSR | Health-Adaptive Time-Scale Representation | HUMS | Health Usage Monitoring System |
| iF | Isolation Forest | IMS | Intelligent Maintenance Systems |
| IRT | Infrared Thermal | JNU | Jiangnan University |
| kNN | K-Nearest Neighbors | LOF | Local Outlier Factor |
| LS-GAN | Least Squares GAN | LS-SVM | Least Square Support Vector Machine |
| LSTM | Long Short-Term Memory | MaFaulDa | Machinery Fault Database |
| MCSA | Motor Current Signature Analysis | MFD | Machinery Fault Diagnosis |
| MFPT | Motor Fault Prognostic Testbed | MFS | Machinery Fault Simulator |
| MFS-RDS | MFS- Rotor Dynamics Simulator | MHM | Machine Health Monitoring |
| ML | Machine Learning | MLP | Multilayer Perceptron |
| ML-TRL | Multi-Label Transfer Reinforcement Learning | MMD | Mean Maximum Discrepancy |
| MWK | Multi-Wavelet Kernel | NAS | Neural Architecture Search |
| NBC | Naive Bayes Classifier | NEU | North East University |
| NREL | National Renewable Energy Laboratory | OC-SVM | One Class Support Vector Machines |
| PCA | Principal Component Analysis | PdM | Predictive Maintenance |
| PHM | Prognostics and Health Management | PINN | Physics Informed Neural Networks |
| PM | Physical Model | PnM | Preventive Maintenance |
| PPO | Proximal Policy Optimization | PSO | Particle Swarm Optimization |
| PU | Paderborn University | RADIS | Real-time Anomaly Detection Intelligent System |
| RBM | Restricted Boltzmann Machines | RF | Random Forest |
| RL | Reinforcement Learning | RM | Reactive Maintenance |
| RNN | Recurrent Neural Network | RPM | Revolutions per Minute |
| RUL | Remaining Useful Life | SCADA | Supervisory Control and Data Acquisition |
| SDU | Shandong University | SEU | Southeast University |
| SL | Supervised Learning | SpAE | Sparse AE |
| SQ | SpectraQuest | SRU | Simple Recurrent Units |
| SSL | Semi-Supervised Learning | StAE | Stacked AE |
| STFT | Short-Time Fourier Transform | SVM | Support Vector Machines |
| TEP | Tennessee Eastman Process | THU | Tsinghua University |
| TL | Transfer Learning | TML | Traditional Machine Learning |
| UA-FS | University of Alberta - Fixed Shaft | UESTC | University of Electronic Science and Technology of China |
| UO | University of Ottawa | UoC | University of Connecticut |
| UORED-VAFCLS | University of Ottawa Rolling-element Dataset – Vibration and Acoustic Faults under Constant Load and Speed Conditions | USL | Unsupervised Learning |
| VAE | Variational AE | WGAN | Wasserstein GAN |
| WK-CNN | Wide-Kernel CNN | WPT | Wavelet Packet Transform |
| WSCN | Weight-Shared Capsule Network | WT | Wavelet Transforms |
| XAI | Explainable Artificial Intelligence | XJTU-SY | Xi'an Jiaotong University and Changxing Sumyoung Technology |

## Appendix B: Machine Fault Detection Datasets

### *Airbus Helicopter Accelerometer Dataset*

Available at ETH Zurich Repository, this dataset includes vibration data from Airbus helicopters, with 1-minute sequences recorded at 1024 Hz. It comprises 1677 training sequences and 594 validation sequences, with half labeled as abnormal. The dataset supports automated flight test data validation and was used in Airbus's 2019 AI challenge to classify sequences as healthy or faulty [33].

### *AI4I 2020 Predictive Maintenance Dataset*

The AI4I 2020 Predictive Maintenance Dataset is a synthetic dataset designed for predictive maintenance in the computer industry. It contains 10,000 data points with 14 features each, such as product ID, air and process temperatures, rotational speed, torque, tool wear, and a 'machine failure' label. The label identifies if a failure has occurred, without specifying the failure mode—tool wear, heat dissipation, power, overstrain, or random. For dataset access, visit this link [34, 35].

### *C-MAPSS Dataset*

The C-MAPSS dataset, derived from the Commercial Modular Aero-Propulsion System Simulation (C-MAPSS), is pivotal for developing predictive models in aerospace fault detection. This dataset is sourced from a dynamic simulator designed for large commercial turbofan engines and includes 21 sensor variables like temperature, pressure, and speed across five subsets depicting various wear levels and degradation processes. It supports testing of advanced algorithms through user-defined transient simulations and linear state-space models. Although ideal for training deep learning models due to its extensive training samples and diverse operating conditions, it's important to note that as a simulated dataset, it may not fully replicate real-world scenarios. The dataset is accessible for download at NASA Data Portal.

***CWRU Bearing Dataset:***

The CWRU bearing dataset, widely utilized for MFD research, comprises experimental data featuring faults like inner race, outer race, and ball faults with varying severities under different load conditions. This dataset provides time-domain vibration signals sampled at two different frequencies–12 KHz and 48 KHz– and collected by sensors at different positions. Data encompasses single-point faults introduced via electro-discharge machining, with fault diameters ranging from 7 to 40 mils, motor loads from 0 to 3 horsepower, and speeds between 1720 to 1797 rpm. Available publicly at Case Western Reserve University's site, files are named by fault type, size, and load, such as `B007_0` for a 7 mil ball fault under no load. This comprehensive dataset is crucial for validating fault diagnosis algorithms under various conditions [43].

***DIRG Bearing Dataset***

The Dynamic and Identification Research Group (DIRG) at Politecnico di Torino's Department of Mechanical and Aerospace Engineering provides a dataset on high-speed aeronautical roller bearings tested above 6000 rpm with two accelerometers positioned at points A1 and A2. This dataset includes two experimental sessions: the first tests various damages on bearing B1, such as localized faults on the inner race ring or a single roller, under varied speeds and loads; the second involves a prolonged test of a single damaged bearing over approximately 330 hours at constant speed and load, recording vibration signals at a sampling rate of 51,200 Hz. More details and the dataset download are available in [22, 99] at this link.

***EDGFD***

The Experimental Dataset for Gear Fault Diagnosis (EDGFD) dataset includes radial vibration signals from a gearbox with helical gears under three conditions: healthy, one chipped tooth, and three worn teeth. It features a helical gear system composed of a 15-teeth pinion and a 110-teeth wheel, operating at a speed ratio of 7.33 and a nominal pinion speed of 1420 RPM. The gear mesh frequency typically measures at 355 Hz but was recorded at 365 Hz under test conditions. Data was captured over a 10-second duration in each test scenario and saved in MATLAB mat-file format. The dataset is publicly accessible at this link. More details are available in the article "Experimental Dataset for Gear Fault Diagnosis" [101].

***FEMTO Bearing Dataset***

The FEMTO bearing dataset, also called as **IEEE 2012 Data Challenge Dataset**, provided by the FEMTO-ST Institute in France, comprises time-domain vibration signals, temperature measurements, and speed data from bearings under both normal and accelerated degradation conditions, collected on the PRONOSTIA experimental platform. This platform is designed to accelerate bearing degradation under controlled conditions for real-time monitoring, using a rotational speed sensor and a force sensor to characterize operating conditions. Vibration signals are recorded every 10 seconds at 25.6 kHz, while temperature is logged at 10 Hz. The dataset includes 17 failure instances under three different conditions, essential for developing and validating prognostic models for bearing RUL prediction [30, 115]. The dataset is available at this link.

***Gearbox Fault Diagnosis Data***

The gearbox fault diagnosis dataset comprises vibration data collected using SpectraQuest's Gearbox Fault Diagnostics Simulator. This dataset was obtained using four vibration sensors placed in four distinct directions, capturing data under various load conditions ranging from 0% to 90%. The dataset consists of two different scenarios: 1) Healthy condition and 2) Broken Tooth Condition. Ten separate text files are available for each case, resulting in a comprehensive collection of data for gearbox fault diagnosis [420]. The information related to dataset can be found at this website. And, the dataset can be downloaded from *data.world*, particularly from this website.

***HUMS Dataset***

The Defence Science and Technology Group (DSTG), Melbourne, Australia also has been conducting data challanges on Health and Usage Monitoring System (HUMS). In 2023, DSTG organized HUMS data challaenge 2023, and the dataset contains vibration data from a planet gear fatigue crack propagation test on a Bell Kiowa 206B-1 helicopter gearbox. The test was conducted at DSTG's Melbourne facility in January 2022, simulating fatigue cracking in helicopter gears at speeds up to 6,000 RPM. The dataset is accessible at HUMS Data Challenge 2023 [421]. Similarly, they have again announced for the data challenge for 2025. This dataset was generated from the Bell 206B-1 Kiowa (OH-58) helicopter main rotor gearbox test program in their helicopter transmission test facility. The link to download the dataset is HUMS Data Challenge 2025.

***IMS Bearing dataset:***

The IMS Bearing dataset from the University of Cincinnati's Center for Intelligent Maintenance Systems is crucial for bearing prognostics. It includes high-resolution time-domain vibration signals and temperature data from run-to-failure tests, captured by accelerometers mounted on test bearings. Data files record one-second vibration snapshots at 20 kHz, facilitating the development and evaluation of data-driven prognostic models for machinery fault detection and prediction [22]. The dataset is accessible from NASA's prognostic data repository or via Kaggle at this link.

***IMTI Dataset***

The Induciton Motor Thermal Image (IMTI) dataset was developed by Najafi et al. [112], featuring 369 thermal images of three-phase induction motors under 11 different fault conditions, captured with a Dali-tech T4/T8 infrared thermal imager at a resolution of 320×240. The dataset, which includes faults like rotor blockage, cooling fan failure, and various stator winding short-circuits, aims to aid in condition monitoring of electrical equipment.

***JNU Bearing dataset***

Jiangnan University (JNU) provided comprehensive bearing datasets for analysis and fault detection. The datasets consist of three vibration datasets, containing one health state and three fault modes which include inner ring fault, outer ring fault, and rolling element fault, recorded at a sampling frequency of 50 kHz, encompassing various rotating speeds. Therefore, the total number of classes was equal to twelve according to different working conditions [115].

***MaFaulDa***

The *MaFaulDa* dataset comprises 1951 multivariate time series collected from a SpectraQuest machinery fault simulator. The dataset represents six simulated states: normal function, imbalance fault, horizontal and vertical misalignment faults, and inner and outer bearing faults. Data were gathered using accelerometers, a tachometer, and a microphone, monitoring two bearing-supported shafts. Faults such as rotor imbalances, shaft misalignments, and defective bearings were systematically introduced. The dataset is available at SMT UFRJ link and Kaggle link.

***MFPT Bearing dataset***

The Motor Fault Pronostic Testbed (MFPT) bearing dataset, provided by the Society for Machinery Failure Prevention Technology, features data from a test rig using a NICE bearing (roller diameter: 0.235, pitch diameter: 1.245, number of elements: 8, contact angle: 0). This dataset captures various fault conditions and baseline measurements, including three baseline conditions, ten fault conditions across outer and inner races under varying loads, and specific data on real-world examples from wind turbine bearings. Data are available *.mat* format and include load, shaft rate, and sample rate details. Accompanied by sample code, this dataset aims to support researchers and CBM practitioners in advancing their techniques. Further information and downloads are available at MFPT website.

***NEU Dataset***

North East University dataset comprises 1,800 steel surface defect images, encompassing six distinct fault types, including crazing, patches, rolled-in scale, inclusion, pitted surface, and scratches. Notably, the dataset maintains a balanced distribution among these fault types, making it a valuable resource for training and testing fault detection models [422]. This dataset can be downloaded from this website.

***NREL Wind Turbine Gearbox Condition Monitoring Vibration Analysis Benchmarking Datasets***

The NREL Wind Turbine Gearbox Condition Monitoring Vibration Analysis Benchmarking Datasets by the National Renewable Energy Laboratory (NREL) address the scarcity of benchmarking datasets for wind turbine CBM systems. This initiative, part of the Gearbox Reliability Collaborative, includes data from both a healthy and a similarly designed damaged gearbox that experienced loss-of-oil events, affecting its bearings and gears [119]. Data from these gearboxes, which underwent dynamometer testing, are available for download here. The test gearboxes, capable of operating at 1800 rpm and 1200 rpm, were outfitted with over 125 sensors, providing extensive data to support research and development in vibration-based condition monitoring techniques. This dataset is crucial for validating and advancing CBM systems in the wind industry.

### *PHM2009 Gearbox Dataset*

The PHM-2009 gearbox dataset, shared by the IEEE international conference on the PHM 2009 data challenge, is valuable for evaluating gearbox fault detection algorithms. This dataset contains 20 test cases with vibration signals, temperature and torque measurements. Pre-processing and feature extraction are required before the dataset can be used for machine learning algorithms. Researchers have developed fault detection algorithms using various traditional and advanced methods and compared their performance using evaluation metrics. The dataset remains an important benchmark for further development of error detection methods [423]. This dataset can be downloaded from this link.

### *PHM2010 Milling Cutter Dataset*

The dataset from the 2010 PHM Society Conference Data Challenge focuses on the RUL estimation for CNC milling machine cutters. Utilizing a Kistler quartz 3-component dynamometer and three accelerometers, it captures cutting forces and vibrations across the X, Y, Z axes, while an acoustic emission sensor monitors high-frequency stress waves during cutting. Operational conditions include a spindle speed of 10400 rpm, a feed rate of 1555 mm/min in the x-direction, and cutting depths of 0.125 mm and 0.2 mm in the y and z directions, respectively, with data sampled at 50 kHz [424]. Despite its collection under a dry milling setup which deviates from typical industrial conditions, the dataset offers valuable insights for RUL prediction in milling operations. It includes 3D cutting force and vibration data along with acoustic signals, supporting both single-sensor and multi-sensor fusion analyses. However, only three of the six cutter datasets are labeled, and the uniform milling condition limits the diversity for cross-prediction scenarios.

### *PHME Datasets*

The European conference on Prognostics and Heath Managements (PHME) has been conducting data challenges. In the year 2021, the challange focused on fault detection, classification, and root cause identification in a manufacturing production line setup. Participants were provided with real-world industrial datasets to apply state-of-the-art algorithms and models. Similarly, participants were tasked with solving a classification problem related to a real production line equipped with Industry 4.0 technologies in the year 2022. The extensive dataset was provided by Bitron Spa, focusing on the operation of automated and integrated machines. Furthermore, the recent data challenge in the year 2024 involved estimating the RUL of a subway ticket validation door system. Participants were encouraged to develop models to predict system failure based on the degradation of door position over time without knowing the operating conditions. All the dataset can be found in their GitHubPage.

### *PU Bearing Dataset*

The PU (Paderborn University) bearing dataset, provided by the KAT data center in Paderborn University includes high-resolution vibration data, torque, and temperature measurements from a custom-built test rig that simulates various bearing faults under different conditions. It comprises data from six healthy bearings and 26 damaged sets, with damage induced both artificially and through accelerated life tests, and is designed to enhance ML algorithm testing. The complete dataset is available for download here.

### *Rotor Fault Dataset*

The dataset was acquired using a rotor lab bench setup consisting of a DC motor connected to an 850mm rotor via bearings and couplings, with two mass disks (75mm diameter) and a GTS3-TG series simulator for data acquisition. Data capture involved two eddy current sensors processing signals for amplification and filtering by a processor before storage on a computer. Four rotor states—normal, unbalanced, misaligned, and rubbing—were recorded at 1200 r/min with a sampling frequency of 2048 Hz and 1-second sample length. Unbalance was simulated with a 2g mass on the disk; misalignment by adjusting the shaft coupling; rubbing by introducing a screw contact with the shaft. The dataset comprises 180 samples across 45 test groups, split into 80 training and 100 testing samples, each processed via wavelet thresholding and stored as a 2-D matrix. The data is available at Mendeley Data.

### *SEU Gear Fault Dataset*

The SEU gear fault dataset from Southeast University, China, is collected using the Drivetrain Dynamics Simulator. It consists of two subdatasets: bearing and gear data under conditions 20-0 and 30-2 for speed-load. It features motor and gearbox vibrations (x, y, z), motor torque, and includes five gear fault types (healthy, chipped, missing tooth, root, surface) and five bearing faults (healthy, inner, outer, combined, rolling element). This dataset is crucial for studying time-frequency distributions and fault diagnosis. Available for download at GitHub [72].

***TEP Simulation Dataset***

The Tennessee Eastman Process (TEP) is a chemical simulation benchmark used for fault diagnosis in continuous processes, involving 22 simulation runs with 52 variables, and 20 fault types. Available at IEEE Dataport.

***THU Gearbox Dataset***

The THU dataset from Tsinghua University comprises vibration data from gearbox gear fault experiments conducted in 2019 using an HS-200 single-stage planetary gearbox. The setup included a motor (29-31 Hz), a planetary gearbox with artificially damaged gears, and two accelerometers mounted on the gearbox case in the x and y directions. It features nine fault types, including a healthy condition and various levels of damage to sun and planetary gears. While the data provides a broad spectrum for fault analysis, it is derived from a controlled environment, which may not fully replicate real-world conditions [40, 134].

***UA-FS Gearbox Dataset***

Generated by the University of Alberta (UA), this dataset contains vibration and speed signals from a fixed-shaft (FS) gearbox, highlighting five levels of tooth root crack severity. Data were collected under two speed profiles with a total of 255,037 samples.

***UO Bearing Dataset***

Provided by the University of Ottawa (UO), this dataset features acceleration and speed signals from bearings under various health conditions. It is available at Mendeley Data.

***UoC Gearbox Fault Dataset***

This dataset is provided by University of Connecticut (UoC). The dataset contains a collection of vibration data from two-stage gears under various operating conditions, and includes data from healthy conditions as well as conditions with various gear defects such as missing teeth, root cracks, spalling and flaking chips. The dataset can be downloaded from the UoC website. The sampling frequency was 20 kHz [40, 425, 426]. The UOC gear failure dataset is a valuable resource for researchers and technicians involved in gear failure diagnosis. It can be used to train and evaluate machine learning models for transmission fault detection. The dataset can also be used to develop new methods for gearbox fault diagnosis.

***UORED-VAFCLS Dataset***

University of Ottawa Rolling-element Dataset – Vibration and Acoustic Faults under Constant Load and Speed conditions (UoARED-VAFCLS) dataset offers vibration and acoustic data for fault analysis under constant load and speed, available at Mendeley Data.

***Wind Turbine SCADA dataset***

The dataset contains supervisory control and data acquisition (SCADA) data collected at wind farms. SCADA systems record various measurements, including wind speed, power output and rotor speed. The dataset helps to analyse wind turbine performance and detect abnormal behaviour, such as gearbox, generator or other component failures [144]. Some of the SCADA datasets are:

(i) Operation SCADA dataset of an urban small wind turbine in São Paulo, Brazil
(ii) 2018 Scada Data of a Wind Turbine in Turkey (Kaggle)
(iii) Wind Turbine SCADA open data (GitHub).

***XJTU-SY Bearing Dataset***

The XJTU-SY bearing dataset is provided by Xi'an Jiaotong University and Changxing Sumyoung Technology Co. It includes run-to-failure data from 15 rolling element bearings under three conditions: 2100 rpm with 12 kN, 2250 rpm with 11 kN, and 2400 rpm with 10 kN. Tests were conducted using an AC motor, a speed controller, a shaft, support bearings, and a hydraulic loader. Data from two accelerometers at 25.6 kHz, capturing 32768 points per minute, are stored in CSV files. This dataset is ideal for validating prognostics algorithms and can be accessed at this link.

### *MFS*

Machine fault simulators (MFS) are critical tools for studying machinery faults in controlled environments, used extensively for over two decades. These simulators integrate robust multi-channel data acquisition systems and comprehensive data analysis software, supporting a variety of analyses such as Time Waveform, Amplitude Spectrum, and Frequency Response [128]. SpectraQuest (SQ) and TIERA are leading providers, offering models like the MFS–Lite, Machinery Fault & Rotor Dynamics Simulator, and MFS–Magnum. SQ's systems are noted for their versatility and ease of use, while TIERA's feature modular designs that facilitate the study of common faults like unbalance and bearing defects without halting production. These tools are invaluable for advancing practical knowledge in machinery diagnosis and fault detection, with datasets like MaFaulDa enhancing research capabilities.

### *Noteworthy Mentions*

Several other datasets have been utilized in MFD, including the HOUDE Dataset [71], Ball Screw Dataset [71], SDU Dataset [77], UNSW Turbine Blade Dataset [90], Qianpeng Company Gearbox Dataset [124], NCEPU Gear Dataset [95], QPZZ-II Dataset [384], and the Locomotive Bearing Dataset [73]. For a comprehensive list of relevant datasets used in MFD, you can visit this GitHub Page.